\documentclass{article}

\usepackage{PRIMEarxiv}
\usepackage{float}
\usepackage[utf8]{inputenc} 
\usepackage[T1]{fontenc}    
\usepackage{hyperref}       
\usepackage{url}            
\usepackage{booktabs}       
\usepackage{amsfonts}       
\usepackage{nicefrac}       
\usepackage{microtype}      
\usepackage{lipsum}
\usepackage{fancyhdr}       
\usepackage{graphicx}       
\usepackage{tabularx}
\usepackage{cite}
\usepackage[numbers]{natbib}
\usepackage{tabularx}
\usepackage{siunitx}
\usepackage{enumitem}
\usepackage{amsmath}
\usepackage[affil-it]{authblk}
\usepackage{makecell}
\usepackage{multirow}
\usepackage{makecell}
\usepackage{tcolorbox}
\usepackage{siunitx} 
\usepackage{url} 
\usepackage{subcaption}
\usepackage[header,title,page]{appendix}

\usepackage[title]{appendix}
\usepackage{booktabs}
\usepackage{tabularx}
\usepackage{pdflscape} 
\usepackage{geometry}
\usepackage{changepage} 
\usepackage{pdflscape}  
\usepackage{booktabs}   
\usepackage{newunicodechar}
\usepackage{hyperref}

\graphicspath{{media/}}     

\title{Leveraging Weighted Syntactic and Semantic Context Assessment Summary (wSSAS) Towards Text Categorization Using LLMs}

\author[1]{Shreeya Verma Kathuria\thanks{\texttt{\{shreeya, nitin, sharookh\}@tellagence.com}}}
\author[1]{Nitin Mayande}
\author[1]{Sharookh Daruwalla}

\author[2]{Nitin Joglekar\thanks{\texttt{nitindra.joglekar@villanova.edu}}}

\author[2]{Charles Weber\thanks{\texttt{webercm@pdx.edu}}}

\affil[1]{Tellagence Inc.}
\affil[2]{Villanova School of Business, Villanova University}
\affil[3]{Maseeh College of Engineering and Computer Science, Portland State University}

\begin{document}
\maketitle

\setcounter{footnote}{0}

\begin{abstract}
The use of Large Language Models (LLMs) for reliable, enterprise-grade analytics such as text categorization is often hindered by the stochastic nature of attention mechanisms and sensitivity to noise that compromise their analytical precision and reproducibility. To address these technical frictions, this paper introduces the Weighted Syntactic and Semantic Context Assessment Summary (wSSAS), a deterministic framework designed to enforce data integrity on large-scale, chaotic datasets. We propose a two-phased validation framework that first organizes raw text into a hierarchical classification structure containing Themes, Stories, and Clusters. It then leverages a Signal-to-Noise Ratio (SNR) to prioritize high-value semantic features, ensuring the model's attention remains focused on the most representative data points. By incorporating this scoring mechanism into a Summary-of-Summaries (SoS) architecture, the framework effectively isolates essential information and mitigates background noise during data aggregation. 

Experimental results using Gemini 2.0 Flash Lite across diverse datasets—including Google Business reviews, Amazon Product reviews, and Goodreads Book reviews—demonstrate that wSSAS significantly improves clustering integrity and categorization accuracy. Our findings indicate that wSSAS reduces categorization entropy and provides a reproducible pathway for  improving LLM based summaries based on a high-precision, deterministic process for large-scale text categorization.

\end{abstract}


\section{Introduction}
The field of text categorization and summarization has fundamentally shifted, evolving from a complex engineering challenge—which historically necessitated extensive feature engineering, massive datasets, and prolonged training \cite{manning_introduction_nodate} —into a core capability powered by Large Language Models (LLMs). This transition is underpinned by the superior semantic understanding of LLMs, enabling zero-shot and few-shot learning—the ability to categorize text with little to no prior training \cite{brown_language_2020}. By replacing rigid, bespoke classifiers with fluid foundational models, LLMs have unlocked applications across high-stakes sectors, ranging from real-time misinformation detection in social media \cite{chen_can_2024} to the precise organization of patient records in clinical healthcare environments \cite{hossain_natural_2023}, \cite{agrawal_large_2022}.

Despite this versatility, the path to enterprise-grade reliability remains obstructed by several technical frictions. Current LLM performance is sensitive to prompt engineering; such that minor syntactic variations in instructions can yield drastically different classification outcomes \cite{zhao_calibrate_2021}. Furthermore, the inherent constraints of In-Context Learning token limits restrict the volume of reference examples a model can process \cite{dong_survey_2024}. At a cognitive level, LLMs struggle with nuanced linguistic phenomena such as irony, intensification, and latent bias \cite{srivastava_beyond_2023}. For organizations, these limitations—compounded by a lack of model interpretability and the scarcity of high-quality annotated data for niche domains—create a significant gap between experimental capability and production-ready accuracy.

\subsection{\textit{The Paradox of LLM Creativity: Why Generative AI underperforms in Data Science }}
The primary obstacle to utilizing LLMs for rigorous data science lies in a fundamental architectural conflict: the paradox of generative creativity. LLMs are, by design, engines of probability. While their underlying mechanics are revolutionary for creative synthesis, they are inherently poorly suited for the rigid, invariant requirements of data analytics. The root of this instability is found in the attention mechanism. In a standard generative configuration, the attention mechanism dictates which tokens the model prioritizes during processing \cite{vaswani_attention_2023}. Because these models are optimized for novelty and fluency, the mechanism may assign disparate weights to the same input tokens across successive runs \cite{bender_dangers_2021}. This stochasticity is an asset for creative tasks, but it represents a significant liability for data science tasks where latent space stability is required to ensure data integrity.

In text categorization, this creative variance manifests as decreased accuracy and poor generalization. Research suggests that the inclusion of irrelevant information within the input context can be damaging to performance, as it forces the model to attend to inconsequential patterns \cite{shi_large_2023}. This creates a signal-to-noise deficit that is particularly acute in modern marketing and commercial datasets, where the sheer volume of data often buries actionable insights under layers of technical friction \cite{kreek_training_2018}

This paper addresses the necessity for a deterministic analytical framework capable of improving LLMs from creative assistants into precise instruments of categorization and summarization. By optimizing the Input Context, we seek to bridge the gap between AI potential and analytical execution \cite{liu_pre-train_2023}. Our inquiry is guided by two pivotal research questions:

\begin{enumerate}
    \item Dynamic Context Improvement: Can categorization accuracy be measurably enhanced by replacing static prompts with dynamically generated, custom-tailored context information for specific requests?
    \item Context-Quality Correlation: Is there a quantifiable relationship between the linguistic quality of provided contextual "hints" and the resulting precision of the categorization?
\end{enumerate}
By identifying, isolating, and removing background noise, we aim to ensure the attention mechanism remains focused exclusively on relevant context, thereby establishing improved LLM-driven data integrity for text categorization and summarization.

\subsection{\textit{Hierarchical Contextual Framework for Analytical Integrity}}
We propose Syntactic \& Semantic Attention Summarization (SSAS) \cite{mayande_syntactic_2024} \cite{mayande_leveraging_2025}, a hierarchical contextual framework that replaces the black box unpredictability of standard LLMs with a structured methodology designed to enforce integrity on chaotic datasets \cite{liu_pre-train_2023}. This approach aligns with the Information Bottleneck (IB) principle, which suggests that an optimal model should compress the input to retain only the information most relevant to the target output \cite{tishby_deep_2015}. The philosophy is operationalized through a specialized two-phase framework:

\begin{enumerate}
    \item Contextual Relevance: The process begins by evaluating the data within its specific context. By identifying information relevancy at a granular level, the system determines which data points are pertinent to the defined problem and which are extraneous.
    \item Noise Reduction and Reliability Improvement: Using the derived context from Phase 1, the system systematically reduces dataset noise. By feeding only refined, relevant context into the LLM, we reduce the variance and significantly improve the consistency and reliability of the output
\end{enumerate}
This methodology refines raw, chaotic data into a reliable and analytical relevant dataset. In parallel, by narrowing the model's focus through derived context, this framework ensures that the refined input consistently yields the same results—addressing, to a large extent, the "stochasticity" problem inherent in generative architectures. This framework mitigates operational complexity for domain experts, enabling a shift in focus from algorithmic calibration toward high-level strategic analysis \cite{davenport_all-ai_2023}

The Weighted SSAS (wSSAS) methodology builds upon the foundational SSAS methodology by introducing a rigorous, data-driven prioritization layer. Technical details of SSAS and wSSAS approaches are elaborated in Section 3.

\section{Related Work}
The emergence of Large Language Models (LLMs) has fundamentally redefined text categorization, shifting the paradigm from supervised feature engineering toward zero-shot and few-shot learning \cite{brown_language_2020} \cite{ma_digeo_2023}. However, as these models move from creative synthesis to enterprise-grade analytics, their inherent instability presents significant challenges. Our work builds upon three primary areas of research: the sensitivity of in-context learning \cite{min_metaicl_2022}, the mechanics of attention-based noise \cite{niu_review_2021} \cite{shi_large_2023}, and hierarchical data summarization.

\subsection{\textit{In-Context Learning and Prompt Instability}}
The efficacy of Large Language Models (LLMs) in zero-shot and few-shot regimes is largely governed by the paradigm of In-Context Learning (ICL). However, despite their sophisticated semantic latent spaces, LLMs exhibit a profound and "notorious" sensitivity to the specificities of the input context. Zhao et al. \cite{zhao_calibrate_2021} characterized this as "prompt instability," demonstrating that stochastic variations—such as the permutation of few-shot examples or minor syntactic shifts in instruction templates—can induce significant fluctuations in classification accuracy. This volatility suggests that the standard attention mechanism often converges on "surface-level" patterns rather than underlying logical structures. 
Furthermore, the architectural constraints of the transformer's context window present a dimensional bottleneck. As noted by Dong et al. \cite{dong_survey_2024}, fixed token limits necessitate a zero-sum trade-off between the depth of individual examples and the breadth of the reference set. In enterprise analytics, where datasets are high-dimensional and noisy, this limitation often leads to "recency bias" or the inclusion of non-representative outliers that confound the model's outcomes.
The wSSAS framework departs from traditional ICL by replacing static, heuristically-derived prompts with a dynamically synthesized context \cite{siledar_one_2024}. By applying a precision-filtering pipeline to the input background, we ensure that the "hints" provided to the model are mathematically optimized for representative signals. This transforms the context from a variable, human-engineered instruction into a stable, feature-engineered instrument, effectively addressing  the inherent stochasticity of the generative process.

\subsection{\textit{Attention Mechanisms and the Signal-to-Noise Challenge}}
The "LLM Paradox" identified in this study—wherein generative fluency inversely correlates with analytical precision—is fundamentally rooted in the transformer’s attention mechanism \cite{vaswani_attention_2023}. While the attention layer excels at global dependency modeling, its probabilistic nature becomes a liability when processing chaotic, non-curated datasets. In these environments, the model often fails to distinguish high-value "signal" from background "noise," leading to a degradation of the latent space stability required for rigorous categorization.
Empirical evidence by \cite{shi_large_2023} suggests that the inclusion of irrelevant information within the input context is more detrimental to model performance than the omission of relevant data. This occurs because extraneous tokens force the attention mechanism to allocate significant weights to inconsequential patterns, effectively "diluting" the focus on salient features. This challenge aligns with the Information Bottleneck (IB) principle \cite{tishby_deep_2015}, which posits that an optimal learning system should maximize the compression of input data while retaining only the information most pertinent to the target output.
The wSSAS methodology operationalizes the IB principle by implementing a pre-inference filtering stage involving data refinement. By calculating a Signal-to-Noise Ratio (SNR), the framework systematically suppresses irrelevant data and outliers before they are ingested by the LLM model. This intervention enforces a deterministic focus, ensuring that the transformer’s limited attention budget is reserved exclusively for contextually dense, representative data points. Consequently, the methodology bridges the gap between the stochasticity of generative architectures and the invariance required for enterprise-grade analytics.

\subsection{\textit{Hierarchical Information Compression and Semantic Alignment}}
Historically, clustering and dimensionality reduction have been the standard tools for organizing large, chaotic datasets into meaningful structures. However, these traditional methods typically treat summarization as a simple, one-dimensional task, failing to account for the complex, multi-layered nature of enterprise data.
Standard Retrieval-Augmented Generation (RAG) and recursive summarization techniques often suffer from "information dilution," where the specific nuances of data points are lost during mid-level aggregation \cite{wang_recursively_2025}. While recent advancements in hierarchical information processing have improved document-level understanding, current models often struggle to reconcile strategic top-down intent \cite{im_hierarchical_2023} with bottom-up empirical evidence \cite{christensen_hierarchical_2014}. Our wSSAS framework addresses this by implementing a dual-flow logic that ensures narrative consistency across three distinct levels: Themes, Stories, and Clusters.
The integration of syntactic alignment (structural hierarchy) \cite{ma_structural_nodate} and semantic alignment (latent meaning) \cite{jurafsky_speech_2014}, \cite{nan_semantic_2016} is a recognized frontier in NLP. While Named Entity Recognition (NER) \cite{roy_recent_2021} and Topic Modeling \cite{jelodar_latent_2019} \cite{xun_topic_2016} provide semantic labels, they do not inherently "weight" the importance of data points based on their representative power within a broader narrative. Our approach draws inspiration from Selective Attention mechanisms in cognitive modeling \cite{desimone_neural_1995}, where non-essential "noise" is suppressed prior to high-level cognitive processing. By utilizing the Summary-of-Summaries (SoS) architecture, wSSAS creates a bounded attention environment that forces the LLM to focus on distilled, sentiment-dense narratives rather than being distracted by the stochastic variance of raw, unweighted text.
Finally, our work builds upon the Information Bottleneck (IB) principle \cite{tishby_deep_2015}, which suggests that an optimal analytical model must compress input to retain only the features most relevant to the target output. While generative AI is optimized for creative novelty, the wSSAS methodology enforces analytical integrity by treating the input context as a precision-engineered feature set. This transforms the LLM from a probabilistic generator into a deterministic instrument, providing a scalable solution to the "black box" unpredictability often cited in current enterprise AI research \cite{davenport_all-ai_2023}.

\begin{table*}[t]
\centering
\renewcommand{\arraystretch}{1.4} 
\begin{tabular}{>{\centering\arraybackslash}p{6.5cm} >{\centering\arraybackslash}p{6.5cm}}
\toprule
\textbf{Syntactic Alignment} & \textbf{Semantic Alignment} \\ \midrule
Defines structural rules governing how words combine into grammatical sentences. Acts as a mechanism for models to understand the structural hierarchy of data. & 
Delves into the meaning of words and sentences. Explores how syntactic structures map onto semantic roles to extract the ``who, what, when, where, and why'' of data. \\ \addlinespace

\textbf{Examples:} & \textbf{Examples:} \\
Coarse-to-Fine Retrieval, Spatial Reasoning Improvements \cite{ranasinghe_learning_2024}, Efficient Tuning for Document Visual QA \cite{lewis_generative_2018} & Word Embeddings \cite{mikolov_efficient_2013}, Named Entity Recognition (NER) \cite{roy_recent_2021}, Topic Modeling \cite{jelodar_latent_2019, xun_topic_2016} \\
\bottomrule
\end{tabular}
\vspace{8pt}
\caption{Syntactic vs. Semantic Alignment}
\label{tab:syntactic_semantic}
\end{table*}


\section{Syntactic \& Semantic Attention Summarization (SSAS)}
The strategic rationale behind our SSAS methodology \cite{mayande_syntactic_2024} is the implementation of a bounded attention mechanism. By pre-processing raw text through synchronized syntactic and semantic filters, we constrain the LLM’s focus to high-signal tokens, effectively performing feature engineering at the prompt level. This ensures that the model recognizes the structural hierarchy of the data before the semantic layer interprets the underlying mood or sentiment following the Compositional Semantics principle \cite{partee_lexical_1995}. Table \ref{tab:syntactic_semantic} contrasts the two foundational pillars of our methodology i.e. Syntactic alignment and Semantic alignment.

By combining these alignments, our methodology creates an accurate, distilled summary of the dataset. This summary functions as a specific input prompt that focuses the LLM's attention mechanism \cite{vaswani_attention_2023} to focus on the provided essential information rather than being distracted by the surrounding noise \cite{shi_large_2023}. This alignment is optimized when applied across a rigorous data hierarchy. 

\subsection{\textit{Hierarchical Data Classification: Themes, Stories, and Clusters}}
\begin{figure}[h]
    \centering
    \includegraphics[width=0.8\textwidth]{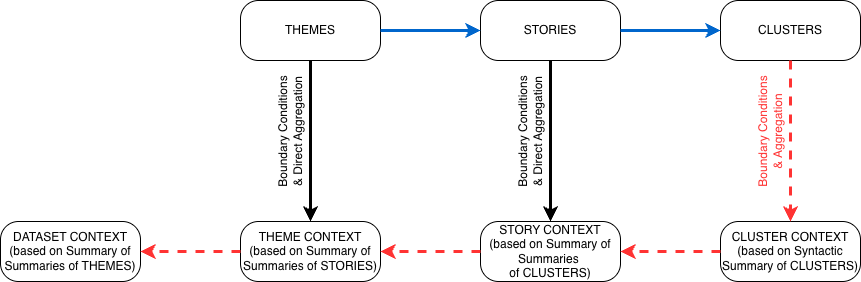}
    \caption{SSAS Architecture for Context Assessment}
    \label{fig:Fig1}
\end{figure}

To transform large-scale, chaotic datasets into actionable insights, a structured hierarchical classification is necessary. Our framework ensures that every data point is evaluated for its contribution to the macro-narrative, preventing the loss of signal in high-volume environments. The SSAS methodology implements three distinct levels found in natural language taxonomies:
\begin{enumerate}
    \item Themes: The most general classification level, identifying the primary macro-topic across all data points within the set.
    \item Stories: The intermediate level of classification, ensuring narrative consistency by identifying specific subtopics within a theme.
    \item Clusters: The lowest level of classification, where the algorithm utilizes localized precision to identify similar data points.
\end{enumerate}
The architecture operates on a dual-flow logic that reconciles strategic intent with empirical evidence, shown in Figure \ref{fig:Fig1}:
\begin{enumerate}
    \item Top-Down Taxonomy (Strategic Intent): The classification flow (Themes -> Stories -> Clusters) organizes data into increasingly granular, manageable segments, a standard approach in Recursive Hierarchy Decoding \cite{im_hierarchical_2023}.
    \item Bottom-Up Aggregation (Data Evidence): The insight flow (Cluster Contexts -> Stories Contexts -> Theme Context) aggregates data to build the Summary of Summaries (SoS), ensuring that high-level insights are grounded in the localized precision of the underlying clusters \cite{christensen_hierarchical_2014}.
\end{enumerate}

\subsection{\textit{Summary-of-Summaries (SoS)}}
The implementation culminates in the context localized Summary-of-Summaries (SoS) architecture. This approach on data pre-processing strategically bounds the LLM's focus by providing a concise, iterative summary as the primary input prompt, effectively reducing the probability of stochastic drift —a phenomenon where the model loses its objective over long context windows \cite{wang_recursively_2025}. The process follows a specific aggregation path:
\begin{enumerate}
    \item Cluster Context: A syntactic summary of individual clusters.
    \item Story Context: An aggregated summary based on the summary of cluster summaries within that story \cite{christensen_hierarchical_2014}.
    \item Theme Context: An aggregated summary based on the summary of story summaries within that theme.
\end{enumerate}

The strategic value of SoS is its role as a feature engineering step. By distilling raw text into a sentiment-dense narrative, we force the LLM to align with core structures and factual content mitigating the risk of "distraction" from irrelevant tokens \cite{li_compressing_2023}

\subsection{\textit{Signal-to-Noise Ratio (SNR) }}
Maintaining data integrity requires a rigorous weighting algorithm to isolate high-value signal from noise. SSAS further uses a weighting logic to validate data points across the hierarchical strata. The primary metric is the Signal to Noise Ratio (SNR), which is the weighted aggregate of three distinct dimensions and is calculated using Equation \eqref{eq:snr_calculation}.

The signal-to-noise ratio ($SNR_i$) is calculated as follows:
\begin{equation}
\label{eq:snr_calculation}
    SNR_{i} = \sum (S_{Theme} + S_{Story}+ S_{Cluster})
\end{equation}
where:
\begin{itemize}
    \item $S_{Theme}$ Theme Signal-to-Noise Ratio that measures global alignment i.e. whether the data point fits the macro-topic
    
    \item $S_{Story}$ Story Signal-to-Noise Ratio that measures narrative consistency i.e. whether the data point fits the sub-topic

    \item $S_{Cluster}$ Cluster Signal-to-Noise Ratio that measures localized precision, i.e., whether the data point fits the immediate group.
    
\end{itemize}

In addition, the methodology incorporates Weighted Amplitude, where keywords are weighted by frequency to enhance the signal. The outcome is Precision Filtering, which suppresses data points that share keywords but lack the contextual depth required for stable attention \cite{shi_large_2023}. This ensures that the LLM is prompted only with high-signal data that perfectly fits the hierarchy, preventing out-of-context noise.

\subsection{\textit{Noise Mitigation: Irrelevant Data and Outlier Management}}
Reliable text categorization necessitates aggressive noise removal to prevent the dilution of the model's focus. SSAS categorizes noise into ranked ordering in two steps:
\begin{enumerate}
    \item Irrelevant Data: This comprises data that does not fit into any defined classification level. Our algorithm labels this as irrelevant data.
    \item Outliers: These are data points within the classification levels that have a negligible impact on the whole level.
\end{enumerate}

Through this rank-ordering, the most representative and contextually dense data points are elevated, while outliers and irrelevant data are suppressed to the bottom of the dataset. This ensures the LLM is prompted only with high-signal data that fits the hierarchy perfectly, preventing out-of-context noise. By dramatically reducing noise in the input data, wSSAS accelerates the identification of core insights, leading to more accurate LLM categorization and superior business decision-making.

\subsection{\textit{Comparison of SSAS and wSSAS Methodology}}

The core difference between SSAS and wSSAS lies in the assignment of analytical value to the derived context summary:
\begin{enumerate}
    \item SSAS (Syntactic \& Semantic Attention Summarization): This unweighted framework provides a structural (syntactic) and meaning-based (semantic) summary, resulting in "Unweighted Context Summary." While it establishes the hierarchical relationships (Themes, Stories, Clusters), it treats all synthesized information as having equal descriptive value. The attention mechanism is bounded by the scope of the summary but is not directed toward the most critical features.

    \item wSSAS (Weighted Syntactic \& Semantic Context Assessment Summarization): This evolved framework applies the calculated Signal-to-Noise Ratio (SNR) to the SSAS-generated summaries, resulting in "Weighted Context Summary." The SNR mathematically prioritizes high-value semantic clusters and narratives, suppressing statistically insignificant data points. This weight layer transforms the context from a complete map of the data (SSAS) into a precision-filtered instrument that actively directs the LLM's attention to the most representative and contextually dense features, effectively isolating "Signal" from "Noise."
\end{enumerate}

\section{Experimental Design: A Two-Phased Validation Framework}
The efficacy of Weighted Syntactic and Semantic Context Assessment Summarization (wSSAS) is validated through a two-phase experimental design. This framework isolates and measures two key components: the quality of the generated context summary and the accuracy of the final categorization, ensuring performance improvements are directly linked to the enhanced input.
\begin{itemize}
    \item Phase 1: Context Summary Quality Assessment: This phase focuses on the algorithmic transformation of raw data. The SSAS algorithm organizes the data into a hierarchy of Themes, Stories, and Clusters. We then generate and compare two distinct context summary types—Unweighted (SSAS) and Weighted (wSSAS)—to determine which offers the most accurate and rich representation of the underlying data signal.
    
    \item Phase 2: Categorization Performance Measurement: This phase evaluates the business impact of the context summary on Large Language Model (LLM) performance.The LLM Gemini 2.0 Flash Lite \cite{gemini_team_gemini_2023} is used to identify primary and secondary topics for each data point. To simplify analysis and enhance interpretability, K-Means clustering is applied to the output, grouping related topics into cohesive category-clusters \footnote{It is important to distinguish Category-Clusters from the Clusters (or Cluster context summary) produced by the SSAS algorithm. Specifically, Category-Clusters are the result of K-Means clustering applied to the Topics that the LLM generates during the categorization phase.}. The experiment compares three scenarios to isolate the wSSAS impact:
    \begin{enumerate}
            \item Baseline: Direct LLM input with no context.
            \item Unweighted Context (SSAS): Categorization using standard SSAS context summary.
            \item Weighted Context (wSSAS): Categorization using the enhanced wSSAS context summary.
    \end{enumerate}

\end{itemize}

\begin{figure}[ht!]
    \centering
    \includegraphics[width=\textwidth]{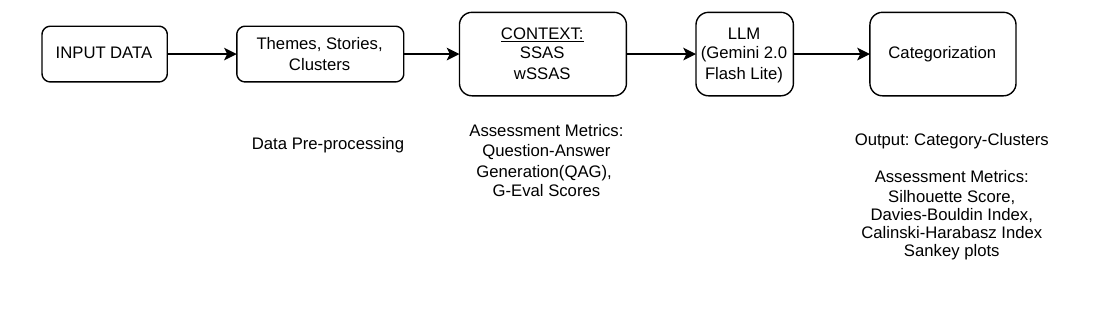}
    \caption{Experimental Design and Assessment Metrics}
    \label{fig:Fig2}
\end{figure}    

As illustrated in Figure \ref{fig:Fig2}, the experimental pipeline moves from raw input data through hierarchical context generation to final categorization (refer Appendix \ref{app:DoE}).The validity of these experiments relies on the rigor of metrics specifically designed to evaluate abstractive intelligence.

\subsection{\textit{Context Summary Evaluation Metrics: QAG and G-Eval}}
Traditional metrics such as ROUGE are insufficient for abstractive summaries as they rely on simple n-gram overlap, failing to capture semantic nuance or factual consistency \cite{lin_rouge_2004}, \cite{liu_g-eval_2023}. We therefore adopt a "LLM-as-a-judge" framework using reference-free metrics.
\begin{enumerate}
\item QAG Mechanics and Embedding Engine: QAG acts as a reference-free "polygraph test" for factual consistency \cite{manakul_selfcheckgpt_2023}. The system generates up to five factual, close-ended questions from the source text and verifies the summary’s ability to provide accurate answers. To calculate semantic similarity between true responses and extracted responses, we utilized the sentence-transformers/all-MiniLM-L6-v2 embedding model \cite{reimers_sentence-bert_2019}
\begin{enumerate}
\item Triage and Encoding: QAG scores were encoded into a 0 (as good as), 1 (better than), or -1 (worse than) scale, comparing weighted vs. unweighted outputs. A critical triage process was applied to prioritize semantic similarity over verbatim alignment. This prevents the penalization of the LLM for utilizing sophisticated paraphrasing while ensuring that factual hallucinations—which an exact-match algorithm might miss—are identified and suppressed.
\end{enumerate}
\item G-Eval Assessment: G-Eval complements QAG by leveraging the LLM to approximate human-like judgment across four qualitative dimensions \cite{liu_g-eval_2023}:
Coherence: Logical structure and organizational flow.
Fluency: Grammatical precision and linguistic naturalism.
Relevance: The concentration of high-value information.
Consistency: Factual alignment with the source manifold.
This approach has been shown to outperform traditional metrics in correlation with human preference.
\end{enumerate}

\subsection{\textit{Categorization Quality Metrics}}
The final evaluation phase focuses on the structural integrity of the generated category-clusters. Internal validation metrics allow us to assess cluster quality without the need for external, human-labeled ground truth \cite{rousseeuw_silhouettes_1987}. Table~\ref{tab:Table2} describes in detail three metrics used in this study to quantify category-cluster quality.

\begin{table}[ht]
    \centering
    \caption{Clustering Evaluation Metrics and Interpretations}
    \label{tab:Table2}
    \begin{tabularx}{\textwidth}{@{} l X X c @{}}
        \toprule
        \textbf{Metric} & \textbf{Description} & \textbf{Interpretation} & \textbf{Goal} \\
        \midrule
        Silhouette Score \cite{rousseeuw_silhouettes_1987} & 
        Measures cohesion vs. separation for each sample. & 
        Range: $[-1, +1]$ \newline +1: Well-separated \newline 0: Overlapping \newline -1: Misassigned\ & 
        Maximize \\
        \addlinespace
        Davies-Bouldin Index \cite{davies_cluster_1979} & 
        Calculates average similarity between clusters. & 
        Lower score indicates better separation and compactness. 0 is the minimum. & 
        Minimize \\
        \addlinespace
        Calinski-Harabasz Index \cite{calinski_dendrite_1974} & 
        Ratio of between-cluster dispersion to within-cluster dispersion. & 
        Higher score indicates dense and well-separated clusters. & 
        Maximize \\
        \bottomrule
    \end{tabularx}
\end{table}

These metrics mathematically confirm whether the wSSAS context summary enables the LLM to identify distinct, compact, and meaningful categories. To ensure generalizability, these metrics were applied across three diverse, industry-standard datasets. 

\begin{table}[ht]
    \centering
    \caption{Dataset Overview}
    \label{tab:Table3}
    \begin{tabularx}{\textwidth}{@{} l r c c l X @{}}
        \toprule
        \textbf{Dataset} & \textbf{\# Reviews} & \textbf{Date Range} & \textbf{Quarters} & \textbf{Primary Entity} & \textbf{Strategic Intent} \\
        \midrule
        Amazon Product & \num{155745} & 01/01/2020 -- 05/23/2023 & 14 & Stores & Product-related \\
        \addlinespace
        Google Business & \num{121826} & 03/01/2009 -- 08/25/2021 & 45 & Book Titles & Restaurant-related \\
        \addlinespace
        Goodreads Book & \num{157407} & 12/07/2006 -- 11/03/2017 & 51 & Restaurants & Literary, subjective \\
        \bottomrule
    \end{tabularx}
\end{table}

\subsection{\textit{Evaluation Datasets: Multi-Domain Selection and Characteristic Analysis}}
To demonstrate the generalizability of the wSSAS methodology, we utilized three diverse, industry-standard datasets from the University of California, San Diego (UCSD) \cite{ni_justifying_2019} (Table~\ref{tab:Table3})
\begin{enumerate}
\item Google Business Reviews (American \& Fast Food restaurants): 121K reviews from North Dakota, used for restaurant sentiment analysis.
\item Amazon Product Reviews (Health \& Personal Care Products): 155K reviews, focused on product discourse over a 3.5-year window.
\item Goodreads Book Reviews (Spoilers): The full 157K dataset, testing the model's ability to handle long-form narrative spoilers.
\end{enumerate}

The datasets showed significant variability in their timelines: Amazon provided the most compressed data (14 quarters), while Goodreads (45 quarters) and Google (51 quarters) offered longer-term data. We characterized the quarterly data using Normalized Volume (High/Low) and Review Distribution, a metric indicating signal stability by tracking the activity of specific sub-topics over time. (Details in Appendix \ref{app:datachar})

\subsection{\textit{Hierarchical Dataset Analysis: Themes, Stories, and Clusters}}
Understanding the data distribution across hierarchical strata (Themes ->  Stories -> Clusters) is critical for identifying how noise removal impacts signal quality. By removing Theme -1 (Irrelevant data) and subsequent outliers, the wSSAS methodology refines the dataset for high-precision categorization. Table~\ref{tab:data_google}, \ref{tab:data_amazon}, and \ref{tab:data_goodreads} show the overall counts of themes, stories, clusters and data points within each of the three datasets. (See Appendix \ref{app:TSC} for count of stories, clusters, data points within each theme in the dataset before and after removal of noisy data.)

\begin{table}[ht]
    \centering
    \caption{Data Processing Statistics across Datasets}
    \label{tab:Table4}

    \begin{subtable}{\textwidth}
        \centering
        \caption{Google Business Reviews}
        \label{tab:data_google}
        \begin{tabularx}{\linewidth}{@{} X c c c r @{}}
            \toprule
            \textbf{Data Stage} & \textbf{Themes} & \textbf{Stories} & \textbf{Clusters} & \textbf{Data points} \\
            \midrule
            All Data & 15 & 113 & \num{8804} & \num{121826} \\
            Without Irrelevant \& Outlier Data & 12 & 54 & 310 & \num{96434} \\
            \bottomrule
        \end{tabularx}
        \vspace{10pt} 
    \end{subtable}

    \begin{subtable}{\textwidth}
        \centering
        \caption{Amazon Product Reviews}
        \label{tab:data_amazon}
        \begin{tabularx}{\linewidth}{@{} X c c c r @{}}
            \toprule
            \textbf{Data Stage} & \textbf{Themes} & \textbf{Stories} & \textbf{Clusters} & \textbf{Data points} \\
            \midrule
            All Data & 15 & 103 & \num{22791} & \num{155745} \\
            Without Irrelevant \& Outlier Data & 14 & 86 & \num{2034} & \num{116102} \\
            \bottomrule
        \end{tabularx}
        \vspace{10pt}
    \end{subtable}

    \begin{subtable}{\textwidth}
        \centering
        \caption{Goodreads Book Reviews}
        \label{tab:data_goodreads}
        \begin{tabularx}{\linewidth}{@{} X c c c r @{}}
            \toprule
            \textbf{Data Stage} & \textbf{Themes} & \textbf{Stories} & \textbf{Clusters} & \textbf{Data points} \\
            \midrule
            All Data & 12 & 80 & \num{22386} & \num{157407} \\
            Without Irrelevant \& Outlier Data & 11 & 66 & \num{1842} & \num{117133} \\
            \bottomrule
        \end{tabularx}
    \end{subtable}
\end{table}

\section{Results}
The transition from a flat data structure to a hierarchical weighting framework is strategically necessary to ensure that the LLM focuses its finite attention mechanism on high-value information. The following results validate the efficacy of wSSAS in distinguishing meaningful semantic "Signal" from interference.

\subsection{\textit{Comparative Performance of Weighted vs. Unweighted Context Summaries}}
To evaluate context summary quality objectively, a reference-free Question-Answer Generation (QAG) framework was implemented using Gemini 2.0 Flash Lite. This method functions as a "polygraph test" for factual consistency, generating close-ended questions from the source data to determine if the generated contexts maintain narrative integrity. A rigorous triage process was applied to these scores to enhance reliability. 
Data indicates that QAG scores from weighted context summary showed a consistent relative improvement post-triage, and these superior QAG scores correlated directly with improved G-Eval metrics. Specifically, weighted context summaries demonstrated higher performance across majority of the G-eval metrics of coherence, fluency, relevance, and consistency, confirming that hierarchical weighting produces a more faithful representation of the information landscape. Table~\ref{tab:Table5} shows an example of the evaluation of the two context summaries generated for a Story within the Google Business Reviews dataset.  (See Appendix \ref{app:QAG} for more examples).

\vspace{20pt}
The wSSAS methodology has been robustly validated across three key industry datasets. Table~\ref{tab:Table6} summarizes the overall QAG performance across the three datasets post-triage showing percentage of Stories and Themes where wSSAS context summary demonstrated equal or superior representation on QAG and G-Eval metrics as compared to SSAS context summary.

\begin{table}[ht!]
    \centering
    \caption{Comparative Analysis of Weighted vs. Unweighted Context Summaries (Google Business Reviews)}
    \label{tab:Table5}
    \small
    \begin{tabularx}{\textwidth}{@{} XX @{}}
        \toprule
        \multicolumn{2}{c}{\textbf{Story ID: 76}} \\
        \midrule
        \textbf{Weighted Context Summary (wSSAS)} & \textbf{Unweighted Context Summary (SSAS)}\\
        \midrule
        \textbf{An Unremarkable Assessment} & \textbf{A Mediocre Experience} \\
        \addlinespace[0.5em]
        The text consistently uses "average" and its variations to describe a subject, emphasizing its lack of
        distinction. It expresses a desire for everything to be average, longing for the ordinary. Data indicates
        a slightly below-average performance for medical doctors, prompting further investigation. A negative
        assessment, described as "average at best," suggests disappointment. A negative dining experience is
        detailed, with tasteless, overcooked food and unsatisfactory service, leading to a recommendation for alternative dining options & The text describes a consistently average, and often disappointing, experience. Medical doctors’ performance is assessed as slightly below average, prompting a desire for everything to be average. A negative dining experience is detailed, with tasteless, overcooked tortellini and inadequate service. The reviewer expresses dissatisfaction with the food quality and service, highlighting the unremarkable nature of the subject being evaluated and suggesting alternatives for better experiences. The overall sentiment conveys a lack of enthusiasm and a longing for the ordinary. \\
        \addlinespace[1em]
        
        \textbf{Quantitative Evaluation:} & \textbf{Quantitative Evaluation:} \\
        \begin{itemize}[leftmargin=*, noitemsep, topsep=0pt]
            \item QAG Pre-Triage: 1/4
            \item QAG Post-Triage: 2/4
            \item G-Eval Scores:
            \begin{itemize}[label=--]
                \item Coherence: \textbf{0.8}
                \item Relevance: \textbf{1.0}
                \item Fluency: 0.9
                \item Consistency: 0.5
            \end{itemize}
        \end{itemize} & 
        \begin{itemize}[leftmargin=*, noitemsep, topsep=0pt]
            \item QAG Pre-Triage: 3/4
            \item QAG Post-Triage: 2/4
            \item G-Eval Scores:
            \begin{itemize}[label=--]
                \item Coherence: 0.4
                \item Relevance: 0.5
                \item Fluency: \textbf{1.0}
                \item Consistency: 0.5
            
            \end{itemize}
        \end{itemize} \\
        \bottomrule
    \end{tabularx}
\end{table}

\vspace{20pt}
  
\begin{figure}[htbp]

     \centering
     \begin{subfigure}[b]{0.48\textwidth}
         \centering
         \includegraphics[width=\textwidth]{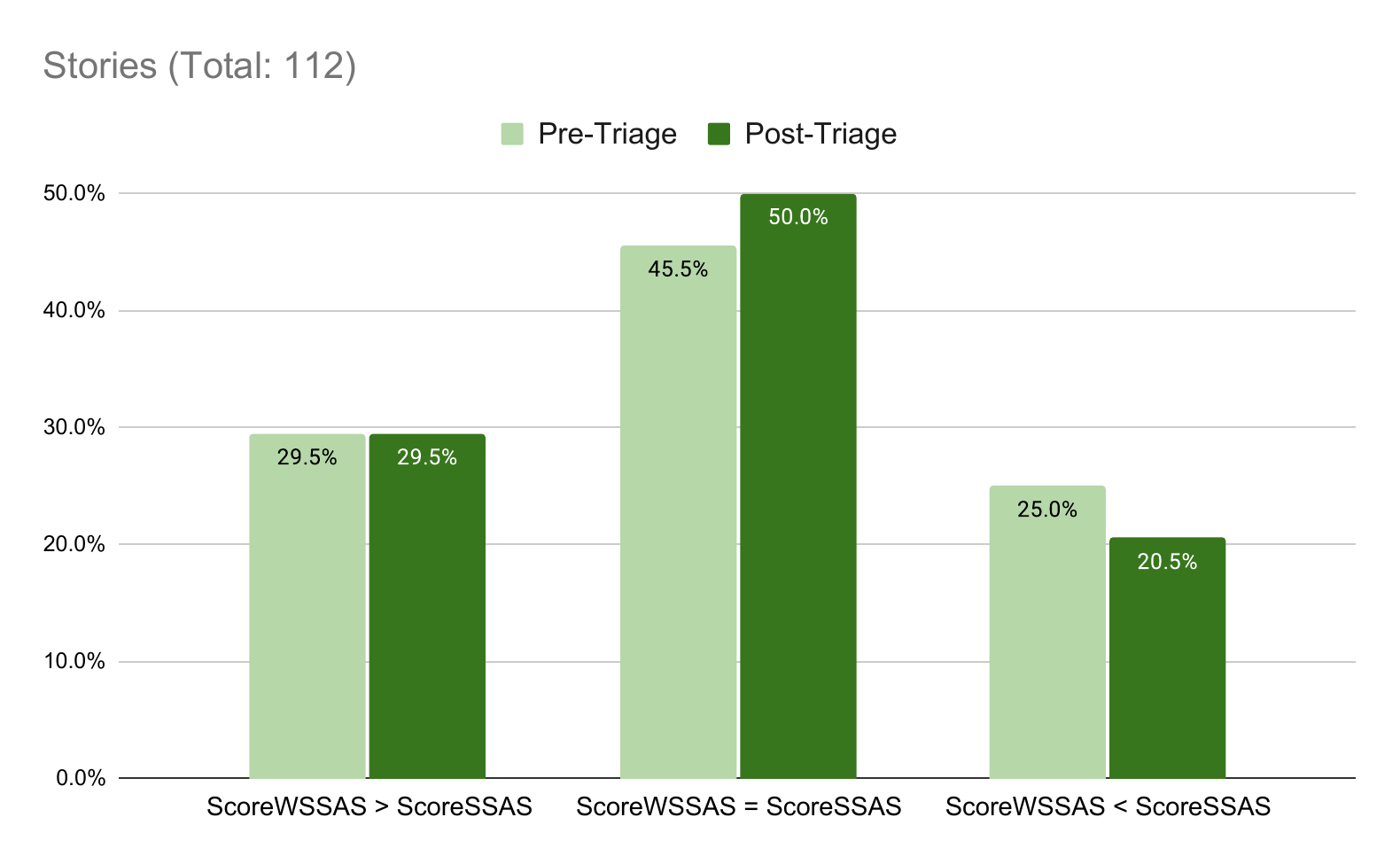}
         \caption{Stories}
         \label{fig:left_side}
     \end{subfigure}
     \hfill 
     \begin{subfigure}[b]{0.48\textwidth}
         \centering
         \includegraphics[width=\textwidth]{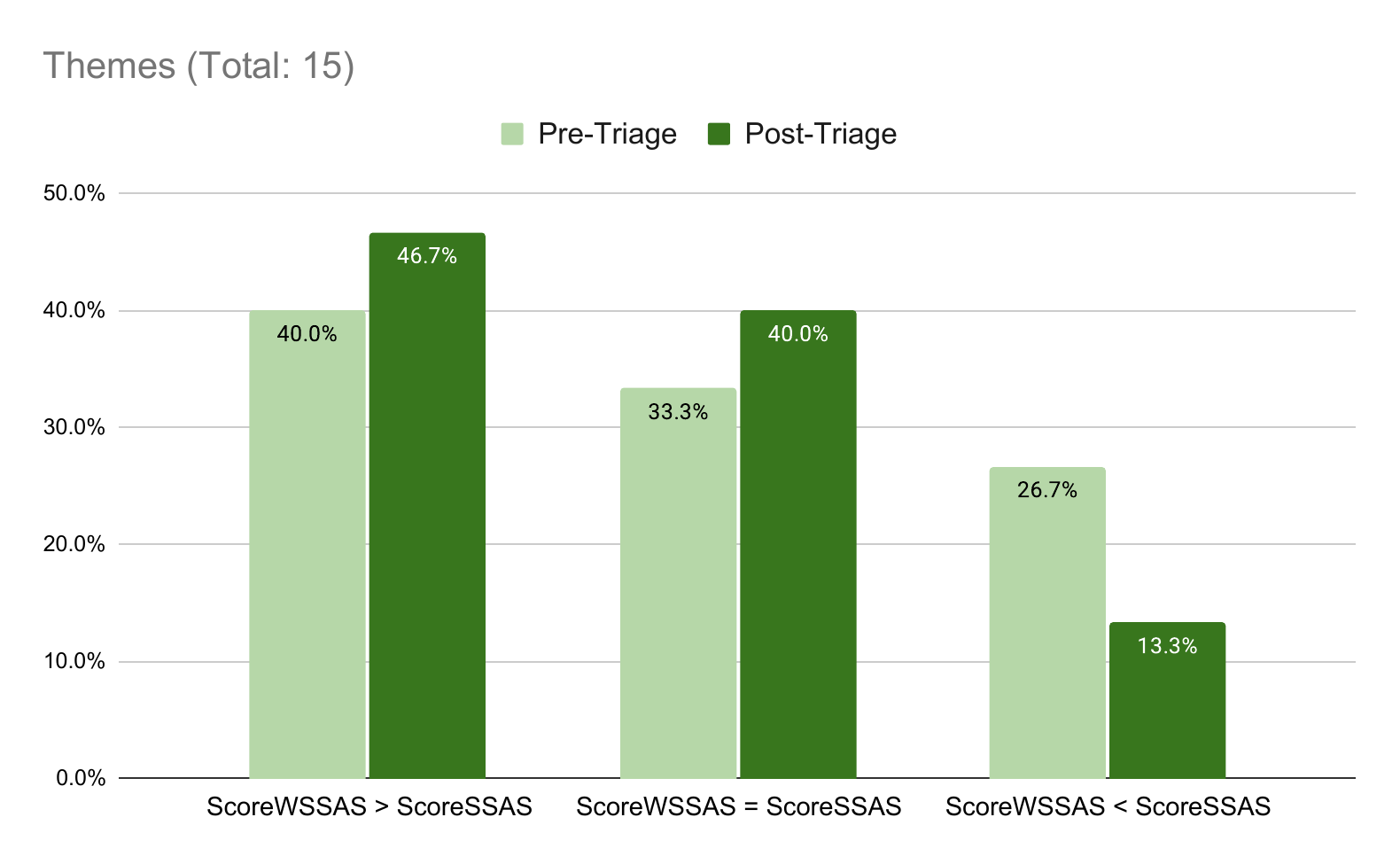}
         \caption{Themes}
         \label{fig:right_side}
     \end{subfigure}
     
\caption{Overall QAG performance for Google Business Reviews}     
\label{fig:Barchart}    
\end{figure}

\vspace{20pt}

\subsection{\textit{Quantitative Assessment of Categorization and Clustering Integrity}}
To validate the quality of the categorization performed, we use three internal validation metrics: the Silhouette Score (measure of cohesion vs. separation), the Davies-Bouldin Index (measure of cluster similarity), and the Calinski-Harabasz (CH) Index (measure of dispersion ratio).
Comparative performance across No Context (Baseline), Unweighted context summary (SSAS), and Weighted context summary(wSSAS) scenarios demonstrates the clear business impact of our contextual grounding approach (Table \ref{tab:clustering_scenarios_all}). The Weighted context approach consistently delivered superior and more actionable clustering across diverse datasets, making it the preferred method for strategic data analysis.

\begin{enumerate}
\item Google Business Reviews: The wSSAS context summary (CH Index: 8006.7) dramatically improved cluster definition and density compared to the "No context" scenario (CH Index: 3041.4), consolidating fragmented data into three strategic categories: "Customer Dissatisfaction \& Service Failures," "Positive Dining Reviews," and "Restaurant Experience and Food Quality. 
\item Amazon Product Reviews: While the SSAS context summary achieved a higher CH Index (4746.3), the superior Silhouette Score (0.049) and Davies-Bouldin Index (3.80) of wSSAS context summary indicate more effective cluster separation and internal cohesion. This suggests that the wSSAS approach provides a better qualitative definition of categories, even if the dispersion ratio is slightly lower than the unweighted model. This is crucial for distinguishing high-density generic feedback from specific, actionable issues like "Defective or Faulty Products."
\item Goodreads Book Reviews: The wSSAS context summary effectively consolidated complex review data into three highly defined clusters, delivering focused, high-density categories such as "Book Reviews and Criticism," "Book Series and Character Relationships," and "Romance and Suspense," preventing the fragmentation seen in the other two scenarios.
\end{enumerate}

Robustness Check: The structural integrity of the wSSAS approach was confirmed; removing irrelevant data points did not materially alter the core thematic architecture, proving the clusters are not easily disrupted by noise or outliers. (See Appendix \ref{app:sankey} for details and Sankey Plot analysis showing the movement of data points between category-clusters for different experimental scenarios)

\begin{table}[ht]
    \centering
    \caption{Overall QAG performance of the wSSAS context summary}
    \label{tab:Table6}
    \begin{tabular}{@{} l S[table-format=2.1] S[table-format=2.1] @{}}
        \toprule
        \textbf{Dataset} & {\textbf{Stories (\%)}} & {\textbf{Themes (\%)}} \\
        \midrule
        Google Business Reviews & 79.5\% & 86.7\% \\
        Amazon Product Reviews  & 87.3\% & 80.0\% \\
        Goodreads Book Reviews  & 86.0\% & 91.7\% \\
        \bottomrule
    \end{tabular}
\end{table}

\begin{table}[H]
    \centering
    \caption{Comparative Clustering Performance and Data Distribution across Categories}
    \label{tab:clustering_scenarios_all}
    \small 

    \begin{subtable}[t]{\textwidth}
        \centering
        \caption{Google Business Reviews}
        \label{tab:results_google}
        \begin{tabular*}{\linewidth}{@{\extracolsep{\fill}} l c p{2.8in} c c c @{}}
            \toprule
            \textbf{Scenario} & \textbf{Count} & \textbf{Category-Cluster Titles (\% Vol)} & 
            \makecell[t]{\textbf{Silhouette}\\\textbf{Score}} & 
            \makecell[t]{\textbf{Davies-Bouldin}\\\textbf{Index}} & 
            \makecell[t]{\textbf{Calinski-Harabasz}\\\textbf{Index}} \\
            \midrule
            \makecell[tl]{Weighted\\Context (wSSAS)} & 3 & 
            \vspace{0pt} 
            \begin{minipage}[t]{2.8in}
            \vspace{-0.85\baselineskip} 
                \begin{itemize}[nosep, leftmargin=*, before=\vspace{-0.6em}, after=\vspace{0.2em}]
                    \item Customer Dissatisfaction (17.7\%)
                    \item Positive Dining Reviews (33.8\%)
                    \item Restaurant Exp. and Food Quality (48.5\%)
                \end{itemize} 
            \end{minipage} & 0.11 & 2.79 & 8006.7 \\
            \addlinespace[12pt]
            \makecell[tl]{Unweighted\\Context (SSAS)} & 3 & 
            \vspace{0pt}
            \begin{itemize}[nosep, leftmargin=*, before=\vspace{-0.6em}, after=\vspace{0.2em}]
                \item Restaurant Customer Satisfaction (26.1\%)
                \item Restaurant Exp. and Operations (59.9\%)
                \item Restaurant Service/Quality (14\%)
            \end{itemize} & 0.07 & 2.92 & 2553.4 \\
            \addlinespace[12pt]
            \makecell[tl]{No context \\ (Baseline)} & 4 & 
            \vspace{0pt}
            \begin{itemize}[nosep, leftmargin=*, before=\vspace{-0.6em}, after=\vspace{0.2em}]
                \item Customer Service/Quality (14.2\%)
                \item Positive Restaurant Experiences (19\%)
                \item Restaurant Exp. and Service (20\%)
                \item Restaurant Reviews/Dining (46.8\%)
            \end{itemize} & 0.06 & 3.35 & 3041.4 \\
            \bottomrule
        \end{tabular*}
    \end{subtable}

    \vspace{10pt}

    \begin{subtable}[t]{\textwidth}
        \centering
        \caption{Amazon Product Reviews}
        \label{tab:results_amazon}
        \begin{tabular*}{\linewidth}{@{\extracolsep{\fill}} l c p{2.8in} c c c @{}}
            \toprule
            \textbf{Scenario} & \textbf{Count} & \textbf{Category-Cluster Titles} & 
            \makecell[t]{\textbf{Silhouette}\\\textbf{Score}} & 
            \makecell[t]{\textbf{Davies-Bouldin}\\\textbf{Index}} & 
            \makecell[t]{\textbf{Calinski-Harabasz}\\\textbf{Index}} \\
            \midrule
            \makecell[tl]{Weighted\\Context (wSSAS)} & 6 & 
            \vspace{0pt}
            \begin{minipage}[t]{2.8in}
            \vspace{-0.85\baselineskip}
                \begin{itemize}[nosep, leftmargin=*, before=\vspace{-0.6em}, after=\vspace{0.2em}]
                    \item Beauty and Grooming Products (15.3\%)
                    \item Cleaning Products (9.7\%)
                    \item Defective/Faulty Products (17.5\%)
                    \item Digestive \& Gut Health Supplements (10.8\%)
                    \item Masks/Accessories (14.1\%)
                    \item Product Reviews/Feedback (32.6\%)
                \end{itemize} 
            \end{minipage} & 0.049 & 3.80 & 4203.0 \\
            \addlinespace[12pt]
            \makecell[tl]{Unweighted\\Context (SSAS)} & 4 & 
            \vspace{0pt}
            \begin{itemize}[nosep, leftmargin=*, before=\vspace{-0.6em}, after=\vspace{0.2em}]
                \item Grooming \& Personal Care (23.4\%)
                \item Pain Relief \& Symptom Management (15.3\%)
                \item Product Defects \& Dissatisfaction (21.8\%)
                \item Product Installation \& User Experience (39.5\%)
            \end{itemize} & 0.047 & 4.13 & 4746.3 \\
            \addlinespace[12pt]
            \makecell[tl]{No context \\ (Baseline)} & 7 & 
            \vspace{0pt}
            \begin{itemize}[nosep, leftmargin=*, before=\vspace{-0.6em}, after=\vspace{0.2em}]
                \item Assistive Devices (11.6\%)
                \item Cleaning \& Maintenance (7.5\%)
                \item Pain \& Symptom Relief (10.2\%)
                \item Personal Grooming \& Hygiene (21.3\%)
                \item Positive Experiences \& Reactions (12.6\%)
                \item Product Functionality \& Performance (11.6\%)
                \item Quality and Performance Issues (25.2\%)
            \end{itemize} & 0.041 & 4.39 & 3264.0 \\
            \bottomrule
        \end{tabular*}
    \end{subtable}

    \vspace{10pt}

    \begin{subtable}[t]{\textwidth}
    \centering
    \caption{Goodreads Book Reviews}
    \label{tab:results_goodreads}
 \begin{tabular*}{\linewidth}{@{\extracolsep{\fill}} l c p{2.8in} c c c @{}}
 \toprule
 \textbf{Scenario} & \textbf{Count} & \textbf{Category-Cluster Titles} &
\makecell[t]{\textbf{Silhouette}\\\textbf{Score}} &
\makecell[t]{\textbf{Davies-Bouldin}\\\textbf{Index}} &
\makecell[t]{\textbf{Calinski-Harabasz}\\\textbf{Index}} \\
\midrule
\makecell[tl]{Weighted\\Context (wSSAS)} & 3 &
\vspace{0pt}
\begin{minipage}[t]{2.8in}
\vspace{-0.85\baselineskip}
\begin{itemize}[nosep, leftmargin=*, before=\vspace{-0.6em}, after=\vspace{0.2em}]
 \item Book Reviews and Criticism (38.5\%)
\item Book Series and Character Relationships (29.4\%)
\item Romance and Suspense (32.1\%)
\end{itemize}
\end{minipage} & 0.041 & 4.61 & 5887.4 \\
\addlinespace[12pt]
\makecell[tl]{Unweighted\\Context (SSAS)} & 5 &
\vspace{0pt}
\begin{itemize}[nosep, leftmargin=*, before=\vspace{-0.6em}, after=\vspace{0.2em}]
\item Book Review Criticism (41.0\%)
\item Book Review Focus (28.3\%)
 \item Character Appreciation Focused Reviews (12.1\%)
\item Content Evaluation and Reaction (7.0\%)
\item Reader Disappointment/Enjoyment (11.7\%)
\end{itemize} & 0.027 & 3.89 & 3453.8 \\
\addlinespace[12pt]
\makecell[tl]{No context \\ (Baseline)} & 3 &
\vspace{0pt}
\begin{itemize}[nosep, leftmargin=*, before=\vspace{-0.6em}, after=\vspace{0.2em}]
\item Book Criticism and Appreciation (28.9\%)
\item Book Review Themes and Tropes (54.4\%)
\item Content Disappointment/Expectation (16.7\%)
\end{itemize} & 0.021 & 4.73 & 5387.7 \\
\bottomrule
\end{tabular*}
\end{subtable}
\end{table}

\section{Conclusion}

\subsection{\textit{Key Research Takeaways}}
The Weighted Syntactic and Semantic Context Assessment Summary (wSSAS) methodology fundamentally reconfigures the data preprocessing landscape by moving beyond the limitations of unweighted architectures. While unweighted models—though providing a complete map of the information landscape—operate under a "flat value structure" that assumes all data points are equal, they ultimately lack the capacity to distinguish critical signals from low-relevance data, resulting in redundant category clusters. In contrast, the wSSAS methodology programmatically engineers a precision-filtered input by utilizing a Signal-to-Noise Ratio (SNR) that validates semantic integrity across three hierarchical strata: Cluster, Story, and Theme signals. This weighted approach ensures that the most representative data rises to the top while mathematically suppressing out-of-context outliers and statistical noise. As confirmed by internal validation metrics, including the Silhouette Score and the Calinski-Harabasz Index, this systematic isolation of high-value semantic signals produces superior, non-redundant category clusters. Ultimately, the primary value of this improved context is its direct contribution to focusing the attention mechanism of Large Language Models (LLMs), providing the high-quality foundation required for hyper-precise categorization.

\subsection{\textit{Impact on Large-Scale Inference}}
\begin{figure}[ht!]
    \centering
    \includegraphics[width=0.5\textwidth]{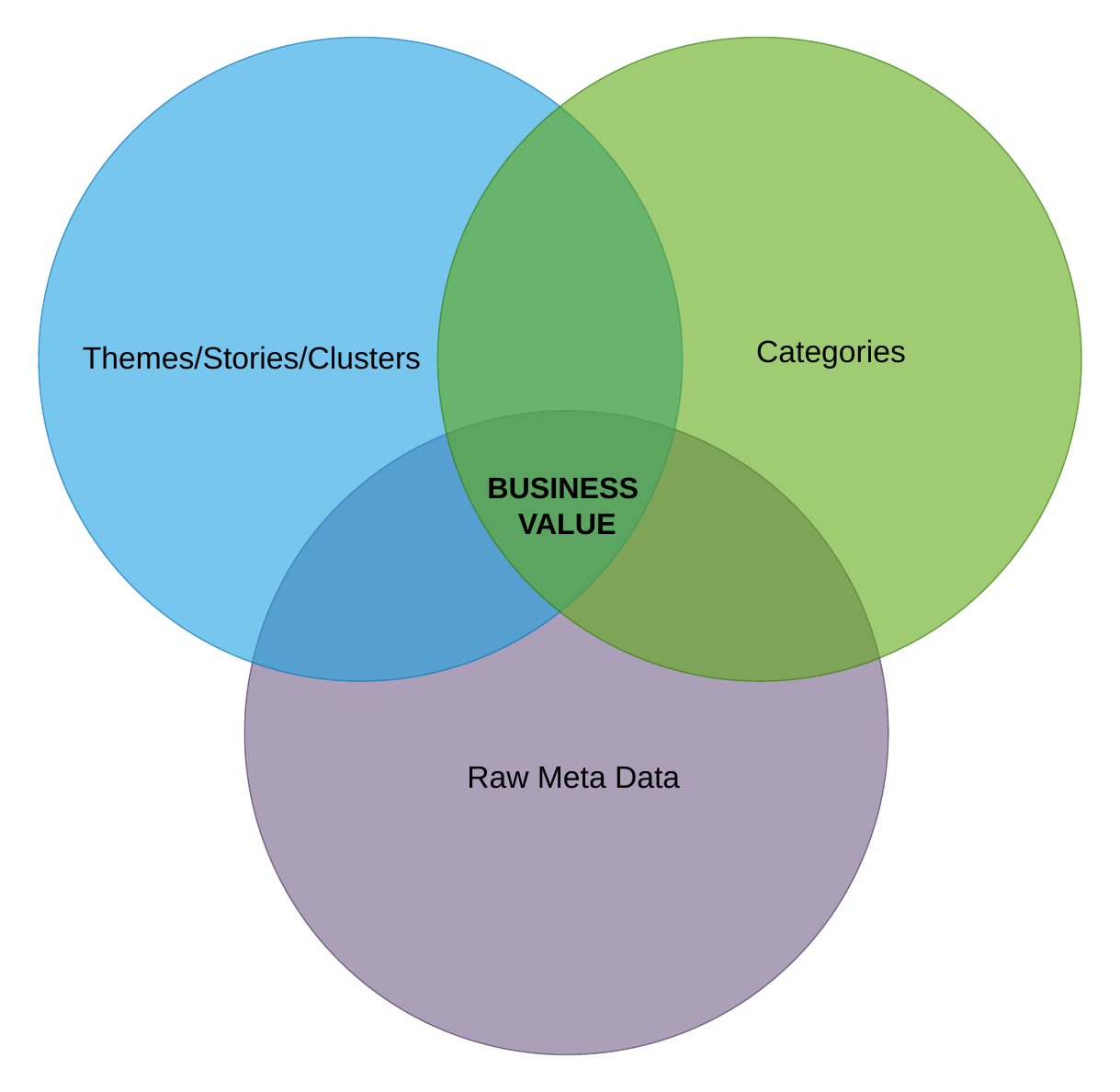}
    \caption{True business value lies at the convergence of generated data-segments
}
    \label{fig:Fig4}
\end{figure}

This methodology's value proposition is its ability to convert heterogeneous data into refined, high-value informational resources, substantially increasing the velocity and precision of organizational decisional processes. By synthesizing hierarchical thematic outputs—comprising Themes, Stories, and Clusters—with automated categorization tools and raw metadata, the framework engineers a unified value proposition. These "derived segments" function as a precision-guided compass for stakeholders, allowing for accelerated  diagnostic and growth  activities across diverse sectors.For instance, restaurant owners can more effectively diagnose the underlying drivers of performance declines, while analysts at a consumer goods firm can leverage these segments to target and acquire new consumer populations with unprecedented accuracy. By converging these high-value components, organizations can extract actionable insights with significantly enhanced speed, ensuring that strategic resources are allocated with maximum efficiency (Figure \ref{fig:Fig4}) To realize these high-level advantages, however, organizations require a well-defined strategy for deploying the wSSAS methodology throughout the enterprise architecture.

\subsection{\textit{Roadmap for Future Work}}
Future research must expand the definition of context into a truly multi-dimensional construct, evolving toward even more granular weighting strata. The next generation of wSSAS will target currently unresolved linguistic ambiguities—specifically complex phenomena such as irony, contrast, and intensification—where models traditionally struggle with reasoning. Beyond linguistic nuances, future iterations should integrate multi-dimensional contextual vectors that account for environmental factors, such as temporal shifts and the quarter-over-quarter trends observed in diverse review datasets. By refining these algorithmic dimensions, the framework will move beyond static summarization toward a dynamic contextual alignment architecture capable of navigating the most intricate intersections of human language and machine intelligence.

\clearpage 
\bibliographystyle{unsrturl}
\bibliography{references_final}  

\clearpage 
\begin{appendices}
\renewcommand{\thetable}{A.\arabic{table}}
\setcounter{table}{0} 

\section{Data Characteristics}
\label{app:datachar}

\begin{table}[ht!]
    \centering
    \label{tab:datachar_all}
    \vspace{1em} 
    \small

    \begin{subtable}{\textwidth}
        \centering
        \caption{Google Business Reviews}
        \label{tab:datachar_google}
        \begin{tabular}{@{} l *{3}{c} @{\hspace{1em}} *{3}{c} @{}}
            \toprule
            & \multicolumn{3}{c}{\textbf{HIGH}} & \multicolumn{3}{c}{\textbf{LOW}} \\
            \cmidrule(r{1em}){2-4} \cmidrule{5-7}
            \textbf{Metric} & \textbf{100} & \textbf{51--99} & \textbf{0--50} & \textbf{100} & \textbf{51--99} & \textbf{0--50} \\
            \midrule
            \# Business & 0 (0\%) & 43 (17.1\%) & 8 (3.2\%) & 0 (0\%) & 32 (12.7\%) & 168 (66.9\%) \\
            \# Datapoints & 0 (0\%) & 93,640 (77\%) & 5,518 (5\%) & 0 (0\%) & 8,734 (7\%) & 13,934 (11\%) \\
            \bottomrule
        \end{tabular}
        \vspace{1.5em}
    \end{subtable}

    \begin{subtable}{\textwidth}
        \centering
        \caption{Amazon Product Reviews}
        \label{tab:datachar_amazon}
        \begin{tabular}{@{} l *{3}{c} @{\hspace{1em}} *{3}{c} @{}}
            \toprule
            & \multicolumn{3}{c}{\textbf{HIGH}} & \multicolumn{3}{c}{\textbf{LOW}} \\
            \cmidrule(r{1em}){2-4} \cmidrule{5-7}
            \textbf{Metric} & \textbf{100} & \textbf{51--99} & \textbf{0--50} & \textbf{100} & \textbf{51--99} & \textbf{0--50} \\
            \midrule
            \# Stores & 87 (0.5\%) & 1,053 (6.4\%) & 1,983 (12.1\%) & 0 (0\%) & 5 (0.0\%) & 13,267 (80.9\%) \\
            \# Datapoints & 17,671 (11\%) & 60,002 (39\%) & 42,503 (27\%) & 0 (0\%) & 44 (0\%) & 35,525 (23\%) \\
            \bottomrule
        \end{tabular}
        \vspace{1.5em}
    \end{subtable}

    \begin{subtable}{\textwidth}
        \centering
        \caption{Goodreads Book Reviews}
        \label{tab:datachar_goodreads}
        \begin{tabular}{@{} l *{3}{c} @{\hspace{1em}} *{3}{c} @{}}
            \toprule
            & \multicolumn{3}{c}{\textbf{HIGH}} & \multicolumn{3}{c}{\textbf{LOW}} \\
            \cmidrule(r{1em}){2-4} \cmidrule{5-7}
            \textbf{Metric} & \textbf{100} & \textbf{51--99} & \textbf{0--50} & \textbf{100} & \textbf{51--99} & \textbf{0--50} \\
            \midrule
            \# Books & 0 (0\%) & 230 (1.2\%) & 4,969 (25.2\%) & 0 (0\%) & 0 (0\%) & 14,526 (73.6\%) \\
            \# Datapoints & 0 (0\%) & 18,720 (12\%) & 93,213 (59\%) & 0 (0\%) & 0 (0\%) & 45,474 (29\%) \\
            \bottomrule
        \end{tabular}
    \end{subtable}

    \bigskip 
    \vspace{0.8em}

  \begin{minipage}{\textwidth}
        Values are presented as Absolute Count (Percentage of Total). 
        \textit{High} and \textit{Low} represent Businesses/Products/Books with Review Volume above the Mean or below the Mean respectively.
        \begin{itemize}
        \item\textit{100}: Reviews present in every available quarter of dataset's lifecycle.
        \item\textit{51-99}: Reviews present in more than half, but not all, quarters
        \item\textit{0-50}: Reviews present in half or fewer quarters
        \end{itemize}

        \vspace{1em}
    \end{minipage}

\end{table}

\clearpage 

\section{Design of Experiments}
\label{app:DoE}
\renewcommand{\thetable}{B.\arabic{table}}

\begin{table}[ht!]
    \centering
    \caption{Experimental Design: Scenarios, Datasets, and Design Rationale}
    \label{tab:experimental_design}
    \small
    \begin{tabularx}{\textwidth}{@{} l l X X @{}}
        \toprule
        \textbf{Scenario} & \textbf{Dataset} & \textbf{Output} & \textbf{Design Rationale} \\
        \midrule
        
        \multirow{3}{*}{\makecell[l]{Base (Themes, \\ Stories, Clusters)}} & Amazon & \multirow{3}{=}{Divided into $N$ Themes; for each theme, we report on \# Stories, \# Clusters, and \# Datapoints} & \multirow{3}{=}{Allows comparison of three industry-standard datasets (ALL, w/o Irrelevant, w/o Outliers) with varying context summaries} \\
        & Google & & \\
        & Goodreads & & \\
        \midrule
        
        \multirow{3}{*}{\makecell[l]{No context \\ (Baseline)}} & Amazon & \multirow{3}{=}{\# Datapoints per category} & \multirow{3}{=}{Categorization using direct LLM input with no context} \\
        & Google & & \\
        & Goodreads & & \\
        \midrule
        
        \multirow{3}{*}{\makecell[l]{Unweighted \\ Context (SSAS)}} & Amazon & \multirow{3}{=}{\# Datapoints per category} & \multirow{3}{=}{Categorization using standard SSAS context} \\
        & Google & & \\
        & Goodreads & & \\
        \midrule
        
        \multirow{3}{*}{\makecell[l]{Weighted \\ Context (wSSAS)}} & Amazon & \multirow{3}{=}{\# Datapoints per category} & \multirow{3}{=}{Categorization using enhanced wSSAS context} \\
        & Google & & \\
        & Goodreads & & \\
        
        \bottomrule
    \end{tabularx}
\end{table}

\clearpage 

\begin{landscape}
\section{Themes, Stories, Clusters breakdown}
\label{app:TSC}
\renewcommand{\thetable}{C.\arabic{table}}
\setcounter{table}{0} 

\begin{table}[ht!]
    \centering
    \caption{Detailed Data Distribution for Google Business Reviews across Noise Removal stages} 
    \label{tab:google_reviews_full_spread}
    
    \small 
    \begin{tabular*}{\linewidth}{@{\extracolsep{\fill}} l rrr @{\hspace{10pt}} l rrr @{\hspace{10pt}} l rrr @{}}
        \toprule
        & \multicolumn{3}{c}{\textbf{All Data}} & \multicolumn{4}{c}{\textbf{W/o Irrelevant Data}} & \multicolumn{4}{c}{\textbf{W/o Irrelevant \& Outlier Data}} \\
        \cmidrule{2-4} \cmidrule{5-8} \cmidrule{9-12}
        
        \textbf{Th.} & \textbf{\#St.} & \textbf{\#Cl.} & \textbf{\#DP} & \textbf{Th.} & \textbf{\#St.} & \textbf{\#Cl.} & \textbf{\#DP} & \textbf{Th.} & \textbf{\#St.} & \textbf{\#Cl.} & \textbf{\#DP} \\
        \midrule
        -1 & 1 & 147 & 273 & & & & & & & & \\
        0 & 11 & \num{3978} & \num{88107} & 0 & 10 & \num{3978} & \num{84789} & 0 & 10 & 190 & \num{74451} \\
        1 & 10 & \num{1146} & \num{10668} & 1 & 9 & \num{1146} & \num{10400} & 1 & 5 & 40 & \num{7631} \\
        2 & 10 & 919 & \num{9299} & 2 & 9 & 919 & \num{9059} & 2 & 7 & 27 & \num{7066} \\
        3 & 10 & 564 & \num{3257} & 3 & 9 & 564 & \num{3184} & 3 & 6 & 9 & \num{1917} \\
        4 & 10 & 435 & \num{2506} & 4 & 9 & 435 & \num{2471} & 4 & 4 & 14 & \num{1386} \\
        5 & 10 & 286 & \num{2145} & 5 & 9 & 286 & \num{2113} & 5 & 7 & 10 & \num{1563} \\
        6 & 10 & 284 & \num{1556} & 6 & 9 & 284 & \num{1531} & 6 & 4 & 5 & 950 \\
        7 & 9 & 311 & 992 & 7 & 8 & 311 & 965 & 7 & 2 & 3 & 230 \\
        8 & 2 & 182 & 842 & 8 & 2 & 182 & 842 & 8 & 1 & 3 & 444 \\
        9 & 11 & 196 & 804 & 9 & 10 & 196 & 803 & 9 & 4 & 4 & 200 \\
        10 & 5 & 70 & 650 & 10 & 5 & 70 & 650 & 10 & 2 & 3 & 486 \\
        11 & 5 & 129 & 370 & 11 & 5 & 129 & 370 & 11 & 2 & 2 & 110 \\
        12 & 5 & 105 & 198 & 12 & 4 & 105 & 196 & & & & \\
        13 & 4 & 52 & 159 & 13 & 3 & 52 & 155 & & & & \\
        \midrule
        \textbf{15} & \textbf{113} & \textbf{\num{8804}} & \textbf{\num{121826}} & \textbf{14} & \textbf{101} & \textbf{\num{8657}} & \textbf{\num{117528}} & \textbf{12} & \textbf{54} & \textbf{310} & \textbf{\num{96434}} \\
        \bottomrule
    \end{tabular*}

    \vspace{10pt}
    \begin{flushleft}
    \footnotesize
    \textit{Note: Th. refers to the Theme ID and \#St, \#Cl, \#DP refer to the number of stories, clusters and data points in the theme. Theme -1 represents unclassified noise. The Subtotal row (Bottom) reflects the number of active themes, total stories, clusters, and datapoints retained in each stage.}
    \end{flushleft}
\end{table}

\begin{table}[ht!]
    \centering
    \caption{Detailed Data Distribution for Amazon Product Reviews across Noise Removal stages} 
    \label{tab:amazon_reviews_full_spread}
    
    \small 
    \begin{tabular*}{\linewidth}{@{\extracolsep{\fill}} l rrr @{\hspace{10pt}} l rrr @{\hspace{10pt}} l rrr @{}}
        \toprule
        & \multicolumn{3}{c}{\textbf{All Data}} & \multicolumn{4}{c}{\textbf{W/o Irrelevant Data}} & \multicolumn{4}{c}{\textbf{W/o Irrelevant \& Outlier Data}} \\
        \cmidrule{2-4} \cmidrule{5-8} \cmidrule{9-12}
        
        \textbf{Th.} & \textbf{\#St.} & \textbf{\#Cl.} & \textbf{\#DP} & \textbf{Th.} & \textbf{\#St.} & \textbf{\#Cl.} & \textbf{\#DP} & \textbf{Th.} & \textbf{\#St.} & \textbf{\#Cl.} & \textbf{\#DP} \\
        \midrule
        -1 & 1 & 344 & 637 & & & & & & & & \\
        0 & 11 & \num{6290} & \num{46974} & 0 & 10 & \num{6290} & \num{44945} & 0 & 10 & 649 & \num{35384} \\
        1 & 10 & \num{4814} & \num{41568} & 1 & 9 & \num{4813} & \num{39507} & 1 & 9 & 404 & \num{32664} \\
        2 & 10 & \num{2339} & \num{15464} & 2 & 9 & \num{2339} & \num{15095} & 2 & 9 & 243 & \num{11456} \\
        3 & 10 & \num{1615} & \num{14701} & 3 & 9 & \num{1615} & \num{14511} & 3 & 9 & 195 & \num{12067} \\
        4 & 10 & \num{2273} & \num{14298} & 4 & 9 & \num{2273} & \num{14033} & 4 & 9 & 204 & \num{10445} \\
        5 & 10 & \num{1602} & \num{10803} & 5 & 9 & \num{1602} & \num{10696} & 5 & 9 & 160 & \num{8234} \\
        6 & 10 & \num{1249} & \num{4204} & 6 & 9 & \num{1249} & \num{4138} & 6 & 8 & 63 & \num{2242} \\
        7 & 10 & 724 & \num{2351} & 7 & 9 & 724 & \num{2163} & 7 & 8 & 34 & \num{1215} \\
        8 & 4 & 668 & \num{2251} & 8 & 3 & 668 & \num{2249} & 8 & 3 & 46 & \num{1200} \\
        9 & 3 & 265 & 803 & 9 & 2 & 265 & 795 & 9 & 2 & 21 & 419 \\
        10 & 2 & 216 & 515 & 10 & 2 & 216 & 515 & 10 & 1 & 5 & 195 \\
        11 & 6 & 215 & 486 & 11 & 6 & 215 & 486 & 11 & 3 & 3 & 148 \\
        12 & 4 & 68 & 396 & 12 & 4 & 68 & 396 & 12 & 4 & 5 & 306 \\
        13 & 2 & 109 & 294 & 13 & 2 & 109 & 294 & 13 & 2 & 2 & 127 \\
        \midrule
        \textbf{15} & \textbf{103} & \textbf{\num{22791}} & \textbf{\num{155745}} & \textbf{14} & \textbf{92} & \textbf{\num{22446}} & \textbf{\num{149823}} & \textbf{14} & \textbf{86} & \textbf{\num{2034}} & \textbf{\num{116102}} \\
        \bottomrule
    \end{tabular*}
    
    \vspace{10pt}
    \begin{flushleft}
    \footnotesize
    \textit{Note: Th. refers to the Theme ID and \#St, \#Cl, \#DP refer to the number of stories, clusters and data points in the theme. Theme -1 represents unclassified noise. The Subtotal row (Bottom) reflects the number of active themes, total stories, clusters, and datapoints retained in each stage.}
    \end{flushleft}
\end{table}

\begin{table}[ht!]
    \centering
    \caption{Detailed Data Distribution for Goodreads Book Reviews across Noise Removal stages} 
    \label{tab:goodreads_reviews_full_spread}
    
    \small 
    \begin{tabular*}{\linewidth}{@{\extracolsep{\fill}} l rrr @{\hspace{10pt}} l rrr @{\hspace{10pt}} l rrr @{}}
        \toprule
        & \multicolumn{3}{c}{\textbf{All Data}} & \multicolumn{4}{c}{\textbf{W/o Irrelevant Data}} & \multicolumn{4}{c}{\textbf{W/o Irrelevant \& Outlier Data}} \\
        \cmidrule{2-4} \cmidrule{5-8} \cmidrule{9-12}
        
        \textbf{Th.} & {\textbf{\#St.}} & {\textbf{\#Cl.}} & \textbf{\#DP} & \textbf{Th.} & \textbf{\#St.} & \textbf{\#Cl.} & \textbf{\#DP} & \textbf{Th.} & \textbf{\#St.} & \textbf{\#Cl.} & \textbf{\#DP} \\
        \midrule
        -1 & 1 & \num{3701} & \num{12443} & & & & & & & & \\
        0 & 11 & \num{11182} & \num{116359} & 0 & 10 & \num{11182} & \num{115048} & 0 & 10 & \num{1349} & \num{99216} \\
        1 & 10 & \num{1143} & \num{4733} & 1 & 9 & \num{1143} & \num{4691} & 1 & 6 & 78 & \num{3081} \\
        2 & 3 & 218 & 605 & 2 & 3 & 218 & 605 & 2 & 3 & 14 & 272 \\
        3 & 10 & \num{1480} & \num{6272} & 3 & 9 & \num{1480} & \num{6167} & 3 & 9 & 117 & \num{4106} \\
        4 & 10 & 905 & \num{2335} & 4 & 9 & 905 & \num{2263} & 4 & 9 & 43 & \num{1031}\\
        5 & 2 & 265 & 714 & 5 & 2 & 265 & 714 & 5 & 2 & 11 & 356 \\
        6 & 7 & 760 & \num{4096} & 6 & 6 & 760 & \num{4095} & 6 & 4 & 55 & \num{3033} \\
        7 & 5 & 451 & \num{1442} & 7 & 5 & 451 & \num{1442} & 7 & 5 & 30 & 832 \\
        8 & 7 & 715 & \num{2296} & 8 & 6 & 715 & \num{2294} & 8 & 6 & 42 & \num{1262} \\
        9 & 10 & 775 & \num{2401} & 9 & 9 & 775 & \num{2370} & 9 & 9 & 46 & \num{1369} \\
        10 & 4 & 791 & \num{3711} & 10 & 4 & 791 & \num{3711} & 10 & 3 & 57 & \num{2575} \\
        \midrule
        \textbf{12} & \textbf{80} & \textbf{\num{22386}} & \textbf{\num{157407}} & \textbf{11} & \textbf{72} & \textbf{\num{18685}} & \textbf{\num{143400}} & \textbf{11} & \textbf{66} & \textbf{\num{1842}} & \textbf{\num{117133}} \\
        \bottomrule
    \end{tabular*}
    
    \vspace{10pt}
    \begin{flushleft}
    \footnotesize
    \textit{Note: Th. refers to the Theme ID and \#St, \#Cl, \#DP refer to the number of stories, clusters and data points in the theme. Theme -1 indicates unclassified noise. The Subtotal row (Bottom) tracks the reduction in active themes and the refinement of cluster density through successive filtering stages.}
    \end{flushleft}
\end{table}

\end{landscape}

\clearpage 
\section{Context Summary Evaluation using QAG / G-Eval Metrics}

\label{app:QAG}
\renewcommand{\thetable}{D.\arabic{table}}
\setcounter{table}{0} 

\begin{table}[ht!]
    \centering
    \caption{Qualitative Comparison of LLM Summarization: Google Business Reviews}
    \label{tab:story_85_comparison}
    \small
    \begin{tabularx}{\textwidth}{@{} XX @{}}
        \toprule
        \multicolumn{2}{c}{\textbf{Story ID: 85}} \\
        \midrule
        \textbf{Weighted Context Summary (wSSAS)} & \textbf{Unweighted Context Summary (SSAS)} \\
        \midrule
        \textbf{Mixed Restaurant Experiences} & \textbf{Mixed Restaurant Experience} \\
        \addlinespace[0.5em]
        The data presents a range of restaurant experiences. Some reviews are overwhelmingly positive, praising food quality and service, with one customer declaring a restaurant their new favorite. Conversely, other reviews highlight negative experiences, including cold service, incorrect orders, and long wait times. One review focuses on the speed and efficiency of a drive-thru. Other reviews express contentment despite a lack of exceptional quality, while another highlights the recurring problem of incomplete orders. One review describes an uncomfortable situation due to a coworker argument. The text also mentions a positive first visit and emphasizes the importance of consistently leaving situations feeling satisfied. Finally, the text describes an environment that is "ridiculously cold" and a persistent state of dissatisfaction. & The text presents a mixed restaurant experience, highlighting both positive and negative aspects. The food, particularly caramel rolls, receives high praise, described as "amazing" and the "best in the world." However, the text also details issues with service, including a cold and inattentive waitress, incorrect orders, and long wait times. The data indicates a recurring problem with incomplete orders, suggesting operational inefficiencies. Despite the food's quality, the inconsistent service and wait times detract from the overall dining experience, creating a trade-off between food quality and service efficiency. \\
        \addlinespace[1em]
        
        \textbf{Quantitative Evaluation:} & \textbf{Quantitative Evaluation:} \\
        \begin{itemize}[leftmargin=*, noitemsep, topsep=0pt]
            \item QAG Pre-Triage: 1/5
            \item QAG Post-Triage: 2/5
            \item G-Eval Scores:
            \begin{itemize}[label=$\circ$, leftmargin=1em]
                \item Coherence: 0.7
                \item Relevance: \textbf{1.0}
                \item Fluency: \textbf{1.0}
                \item Consistency: 0.5
            \end{itemize}
        \end{itemize} & 
        \begin{itemize}[leftmargin=*, noitemsep, topsep=0pt]
            \item QAG Pre-Triage: 3/5
            \item QAG Post-Triage: 2/5
            \item G-Eval Scores:
            \begin{itemize}[label=$\circ$, leftmargin=1em]
                \item Coherence: \textbf{0.8}
                \item Relevance: 0.8
                \item Fluency: 0.9
                \item Consistency: 0.5
            \end{itemize}
        \end{itemize} \\
        \bottomrule
    \end{tabularx}
\end{table}

\begin{table}[ht!]
    \centering
    \caption{Qualitative Comparison of LLM Summarization: Amazon Product Reviews}
    \label{tab:comparison_amazon_39}
    \small
    \begin{tabularx}{\textwidth}{@{} XX @{}}
        \toprule
        \multicolumn{2}{c}{\textbf{Story ID: 39}} \\
        \midrule
        \textbf{Weighted Context Summary (wSSAS)} & \textbf{Unweighted Context Summary (SSAS)} \\
        \midrule
        \textbf{Product Review Summaries} & \textbf{Pain Relief Experiences: A Diverse Overview} \\
        \addlinespace[0.5em]
        The text presents a collection of product reviews, primarily focusing on pain relief. Many reviews are overwhelmingly positive, highlighting effective relief from various conditions like hand pain, arthritis, and back pain, with some users experiencing almost immediate results. Products mentioned include pain relief patches, neck traction devices, and topical applications. Conversely, some reviews report negative experiences, citing ineffectiveness, functional issues, and a lack of relief. The reviews also touch on the ease of use, convenience, and value of the products, with some users expressing a desire to repurchase. & The data presents a wide array of experiences with pain relief products and methods. Positive outcomes are reported for IcyHot, neck traction devices, arthritis treatments, horse massages, red light belts, pain relief patches, and various sprays and creams, with many users experiencing significant and lasting relief. Conversely, negative experiences are documented with ineffective products for finger pain, runner's toe, neck issues, and back pain, along with issues like premature shoe failure and incorrect orders. Some products offer temporary relief, while others provide more sustained benefits. The effectiveness of each method varies, highlighting the diverse nature of pain management and the importance of finding the right solution.

 \\
        \addlinespace[1em]
        
        \textbf{Quantitative Evaluation:} & \textbf{Quantitative Evaluation:} \\
        \begin{itemize}[leftmargin=*, noitemsep, topsep=2pt]
            \item QAG Pre-Triage: 2 / 5
            \item QAG Post-Triage: 3 / 5
            \item G-Eval Scores:
            \begin{itemize}[label=$\circ$, leftmargin=1em]
                \item Coherence: \textbf{0.8}
                \item Relevance: 1.0
                \item Fluency: 0.9
                \item Consistency: 0.5
            \end{itemize}
        \end{itemize} & 
        \begin{itemize}[leftmargin=*, noitemsep, topsep=2pt]
            \item QAG Pre-Triage: 3 / 5
            \item QAG Post-Triage: 3 / 5
            \item G-Eval Scores:
            \begin{itemize}[label=$\circ$, leftmargin=1em]
                \item Coherence: \textbf{0.1}
                \item Relevance: 1.0
                \item Fluency: 1.0
                \item Consistency: 0.5
            \end{itemize}
        \end{itemize} \\
        \bottomrule
    \end{tabularx}
    \vspace{0.5em}
\end{table}

\begin{table}[ht!]
    \centering
    \label{tab:comparison_amazon_47}
    \small
    \begin{tabularx}{\textwidth}{@{} XX @{}}
        \toprule
        \multicolumn{2}{c}{\textbf{Story ID: 47}} \\
        \midrule
        \textbf{Weighted Context Summary (wSSAS)} & \textbf{Unweighted Context Summary (SSAS)} \\
        \midrule
        \textbf{Shoe Horn Reviews: A Mixed Bag} & \textbf{Shoe Horn Reviews: Functionality and Design} \\
        \addlinespace[0.5em]
        The text presents a collection of reviews for shoe horns, revealing a wide range of experiences. Many users praise the telescoping shoe horns for their ease of use, especially for those with mobility issues, highlighting their sturdiness, adjustability, and travel-friendly design. However, other reviews detail significant design flaws, including flimsy construction, ineffective locking mechanisms, and handles that detach. Some users found the shoe horns helpful after surgery or for everyday use, while others considered them a waste of money due to their poor functionality.
         & The text compiles numerous reviews of shoe horns, highlighting both positive and negative experiences. Many users praise telescoping shoe horns for their sturdiness, ease of use, and ability to assist individuals with mobility issues, particularly those with back problems, arthritis, or recovering from surgery. These reviews often emphasize the product's adjustability and portability. Conversely, several reviews criticize shoe horns for design flaws, such as flimsy materials, unstable handles, and ineffective spring mechanisms, rendering them difficult or impossible to use. Some users also report issues with durability, with parts breaking shortly after purchase. \\
        \addlinespace[1em]
        
        \textbf{Quantitative Evaluation:} & \textbf{Quantitative Evaluation:} \\
        \begin{itemize}[leftmargin=*, noitemsep, topsep=2pt]
            \item QAG Pre-Triage: 1 / 5
            \item QAG Post-Triage: 4 / 5
            \item G-Eval Scores:
            \begin{itemize}[label=$\circ$, leftmargin=1em]
                \item Coherence: \textbf{0.9}
                \item Relevance: 1.0
                \item Fluency: 1.0
                \item Consistency: 0.5
            \end{itemize}
        \end{itemize} & 
        \begin{itemize}[leftmargin=*, noitemsep, topsep=2pt]
            \item QAG Pre-Triage: 2 / 5
            \item QAG Post-Triage: 4 / 5
            \item G-Eval Scores:
            \begin{itemize}[label=$\circ$, leftmargin=1em]
                \item Coherence: \textbf{0.8}
                \item Relevance: 1.0
                \item Fluency: 0.9
                \item Consistency: 0.5
            \end{itemize}
        \end{itemize} \\
        \bottomrule
    \end{tabularx}
\end{table}

\begin{table}[ht!]
    \centering
    \caption{Qualitative Comparison of LLM Summarization: Goodreads Book Reviews}
    \label{tab:comparison_goodreads_31}
    \small
    \begin{tabularx}{\textwidth}{@{} XX @{}}
        \toprule
        \multicolumn{2}{c}{\textbf{Story ID: 31 }} \\
        \midrule
        \textbf{Weighted Context Summary (wSSAS)} & \textbf{Unweighted Context Summary (SSAS)} \\
        \midrule
        \textbf{Book Review Roundup} & \textbf{Book Review Roundup} \\
        \addlinespace[0.5em]
        The text summarizes a collection of book reviews, revealing diverse reader experiences. Some reviewers praised engaging plots, well-developed characters, and unexpected twists, while others expressed disappointment with pacing, predictability, or unmet expectations. Reviews covered various genres, with some readers eager for sequels and others expressing mixed feelings, highlighting the subjective nature of reading.& The text summarizes a collection of book reviews, encompassing diverse genres and reader experiences. Reviews range from enthusiastic praise, highlighting engaging plots, well-developed characters, and unexpected twists, to critical assessments, citing predictable storylines, underdeveloped characters, and unmet expectations. The reviews reflect the subjective nature of reading, with some readers finding books deeply moving and others disappointed by various aspects of the narratives. The text also includes reviews of specific books, such as a negative review of a book about the Zodiac Killer, a positive review of a book about a family with secrets, and a mixed review of a book about a romance. \\
        \addlinespace[1em]
        
        \textbf{Quantitative Evaluation:} & \textbf{Quantitative Evaluation:} \\
        \begin{itemize}[leftmargin=*, noitemsep, topsep=2pt]
            \item QAG Pre-Triage: 1 / 5
            \item QAG Post-Triage: 3 / 5
            \item G-Eval Scores:
            \begin{itemize}[label=$\circ$, leftmargin=1em]
                \item Coherence: 1.0
                \item Relevance: \textbf{1.0}
                \item Fluency: 1.0
                \item Consistency: 0.5
            \end{itemize}
        \end{itemize} & 
        \begin{itemize}[leftmargin=*, noitemsep, topsep=2pt]
            \item QAG Pre-Triage: 2 / 5
            \item QAG Post-Triage: 2 / 5
            \item G-Eval Scores:
            \begin{itemize}[label=$\circ$, leftmargin=1em]
                \item Coherence: 1.0
                \item Relevance: \textbf{0.5}
                \item Fluency: 1.0
                \item Consistency: 0.5
            \end{itemize}
        \end{itemize} \\
        \bottomrule
    \end{tabularx}
\end{table}

\begin{table}[ht!]
    \centering
    \label{tab:comparison_goodreads_10}
    \small
    \begin{tabularx}{\textwidth}{@{} XX @{}}
        \toprule
        \multicolumn{2}{c}{\textbf{Theme ID: 10 }} \\
        \midrule
        \textbf{Weighted Context Summary(wSSAS)} & \textbf{Unweighted Context Summary (SSAS)} \\
        \midrule
        \textbf{Diverse Reader Experiences in Book Reviews} & \textbf{Diverse Reader Reactions to Books} \\
        \addlinespace[0.5em]
        The text summarizes a collection of book reviews, revealing a wide spectrum of reader opinions. Some reviews express strong enjoyment, praising engaging plots, relatable characters, and unique settings, while others express disappointment, citing uninteresting characters, predictable plots, and writing styles that failed to resonate. The reviews highlight the subjective nature of reading, with readers responding differently to similar elements like romance, world-building, and character dynamics, reflecting varied preferences and critiques across different genres. & The text summarizes a collection of book reviews, revealing a wide spectrum of reader opinions. Some reviews express strong positive sentiments, praising engaging plots, well-developed characters, and unique premises, while others criticize pacing, character development, and unsatisfying endings. The reviews highlight the subjective nature of reading, with some readers finding books captivating and others disappointed by various elements. Overall, the data reflects a range of opinions across different genres, indicating varied reader preferences and levels of satisfaction. \\
        \addlinespace[1em]
        
        \textbf{Quantitative Evaluation:} & \textbf{Quantitative Evaluation:} \\
        \begin{itemize}[leftmargin=*, noitemsep, topsep=2pt]
            \item QAG Pre-Triage: 3 / 5
            \item QAG Post-Triage: 5 / 5
            \item G-Eval Scores:
            \begin{itemize}[label=$\circ$, leftmargin=1em]
                \item Coherence: 0.8
                \item Relevance: \textbf{1.0}
                \item Fluency: \textbf{1.0}
                \item Consistency: 0.5
            \end{itemize}
        \end{itemize} & 
        \begin{itemize}[leftmargin=*, noitemsep, topsep=2pt]
            \item QAG Pre-Triage: 5 / 5
            \item QAG Post-Triage: 5 / 5
            \item G-Eval Scores:
            \begin{itemize}[label=$\circ$, leftmargin=1em]
                \item Coherence: 0.8
                \item Relevance: \textbf{0.8}
                \item Fluency: \textbf{0.8}
                \item Consistency: 0.5
            \end{itemize}
        \end{itemize} \\
        \bottomrule
    \end{tabularx}
\end{table}

\clearpage
\section{Sankey Plots}
Google Business Reviews (All Data)
\label{app:sankey}

\begin{figure}[ht!]
    \centering
    
    \begin{subfigure}{\textwidth}
    \caption{No Context (Baseline) vs Weighted Context Summary (wSSAS)}
        \centering
        \includegraphics[trim={0 0 0 35pt}, clip, width=0.9\linewidth]{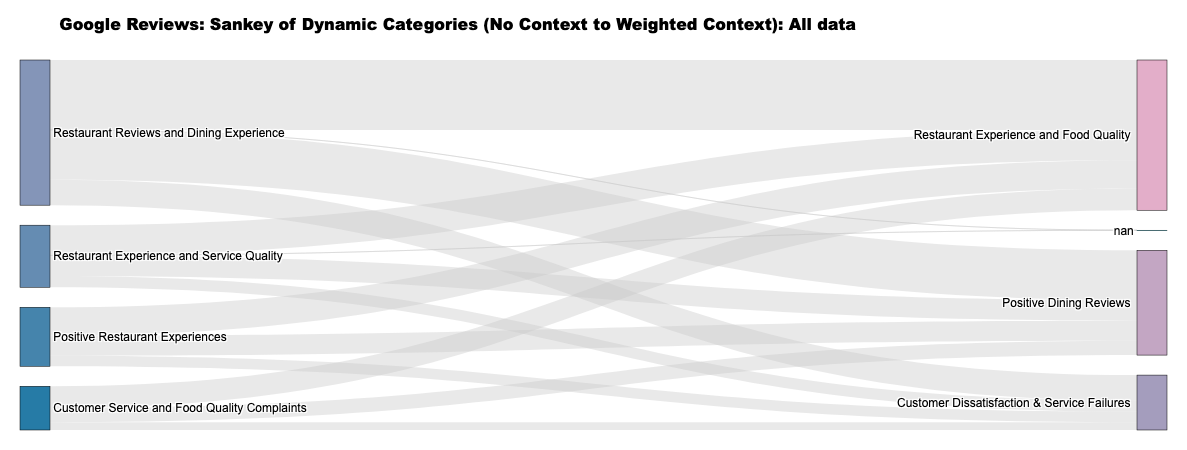}
        
    \end{subfigure}
    
    \vspace{10pt}

    \begin{subfigure}{\textwidth}
    \caption{No Context (Baseline) vs Unweighted Context Summary (SSAS)}
        \centering
        \includegraphics[trim={0 0 0 35pt}, clip, width=0.9\linewidth]{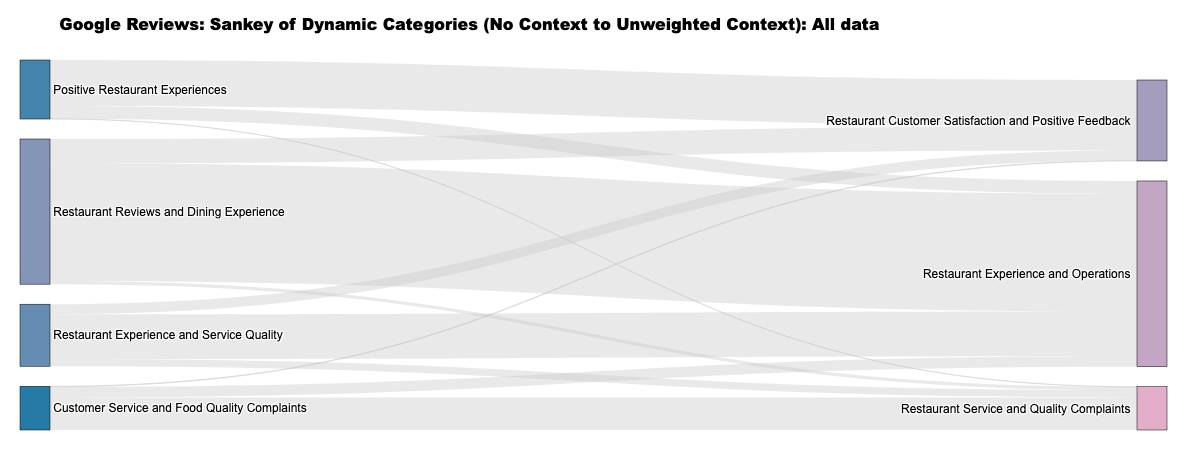}
        
    \end{subfigure}

    \vspace{10pt}

    \begin{subfigure}{\textwidth}
    \caption{Unweighted Context Summary (SSAS) vs Weighted Context Summary (wSSAS)}
        \centering
        \includegraphics[trim={0 0 0 35pt}, clip, width=0.9\linewidth]{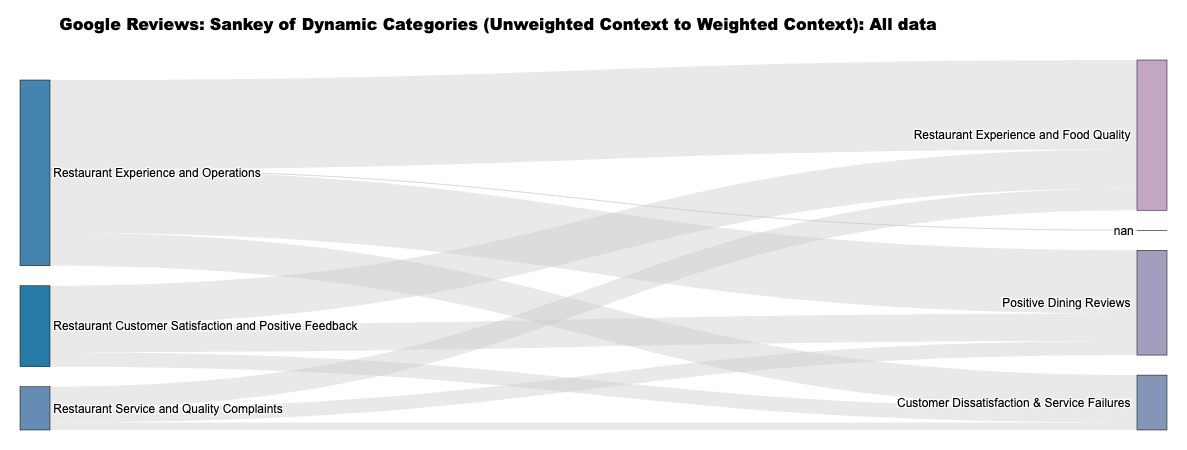}
        
    \end{subfigure}

    \caption{Detailed Sankey diagrams showing cluster transitions for Google Business Reviews.}
\end{figure}

\renewcommand{\thetable}{E.\arabic{table}}
\setcounter{table}{0} 

\begin{table}[ht!]
    \centering
    \caption{Distribution of Review Categories across Experimental Scenarios (Google Business Reviews)}
    \label{tab:category_distribution_full}
    \normalsize 
    \renewcommand{\arraystretch}{1.2} 
    
    \begin{tabularx}{\linewidth}{@{} l >{\raggedright\arraybackslash}X r r r @{}}
        \toprule
        \textbf{Scenario} & \textbf{Category-Cluster Titles} & \textbf{All Data} & \textbf{\makecell[r]{W/o\\Irrelevant}} & \textbf{\makecell[r]{W/o Irrelevant\\\& Outliers}} \\
        \midrule
        
        \multirow{4}{*}{\makecell[l]{No context \\ (Baseline)}} 
        & Customer Service and Food Quality Complaints & 17,189 & 16,566 & 13,302 \\
        & Restaurant Customer Satisfaction and Positive Feedback & 31,818 & 31,116 & 27,991 \\
        & Restaurant Experience and Service Quality & 24,392 & 23,757 & 18,922 \\
        & Restaurant Reviews and Dining Experience & 57,057 & 54,539 & 43,972 \\
        
        \midrule
        
        \multirow{4}{*}{\makecell[l]{Unweighted \\ Context (SSAS)}}
        & Customer Dissat. \& Service Failures & 21,578 & 20,786 & 16,443 \\
        & Positive Restaurant Experiences & 23,188 & 22,666 & 20,238 \\
        & Restaurant Experience and Operations & 73,002 & 70,074 & 55,766 \\
        & Restaurant Service and Quality Complaints & 17,006 & 16,338 & 12,677 \\
        
        \midrule
        
        \multirow{3}{*}{\makecell[l]{Weighted \\ Context (wSSAS)}}
        & Positive Dining Reviews & 41,197 & 40,237 & 35,958 \\
        & Restaurant Experience and Food Quality & 59,044 & 56,498 & 44,028 \\
        & Customer Dissat. \& Service Failures & 21,578 & 20,786 & 16,443 \\ \addlinespace[4pt] 
        
        \bottomrule
    \end{tabularx}
\end{table}

\clearpage
Amazon Product Reviews (All Data)

\begin{figure}[ht!]
    \centering
    
    \begin{subfigure}{\textwidth}
    \caption{No Context (Baseline) vs Weighted Context Summary (wSSAS)}
        \centering
        \includegraphics[trim={0 0 0 35pt}, clip, width=0.9\linewidth]{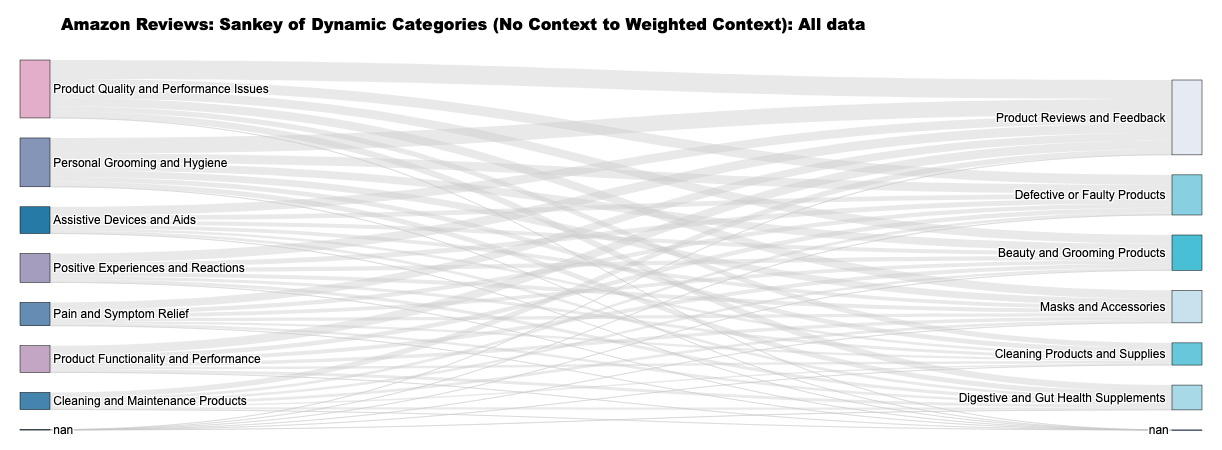}
        
    \end{subfigure}
    
    \vspace{10pt}

    \begin{subfigure}{\textwidth}
    \caption{No Context (Baseline) vs Unweighted Context Summary(SSAS)}
        \centering
        \includegraphics[trim={0 0 0 35pt}, clip, width=0.9\linewidth]{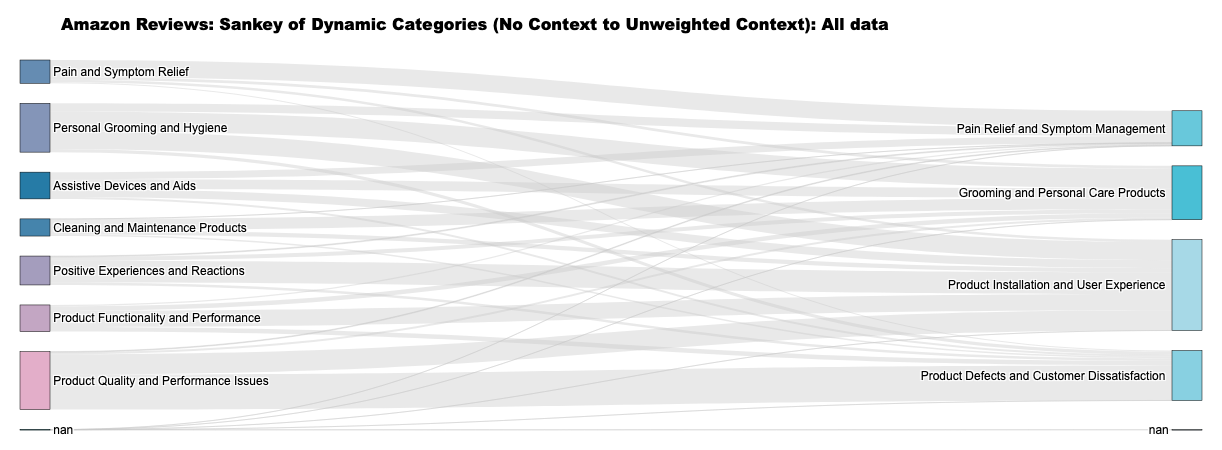}
        
    \end{subfigure}

    \vspace{10pt}

    \begin{subfigure}{\textwidth}
    \caption{Unweighted Context Summary (SSAS) vs Weighted Context Summary (wSSAS)}
        \centering
        \includegraphics[trim={0 0 0 35pt}, clip, width=0.9\linewidth]{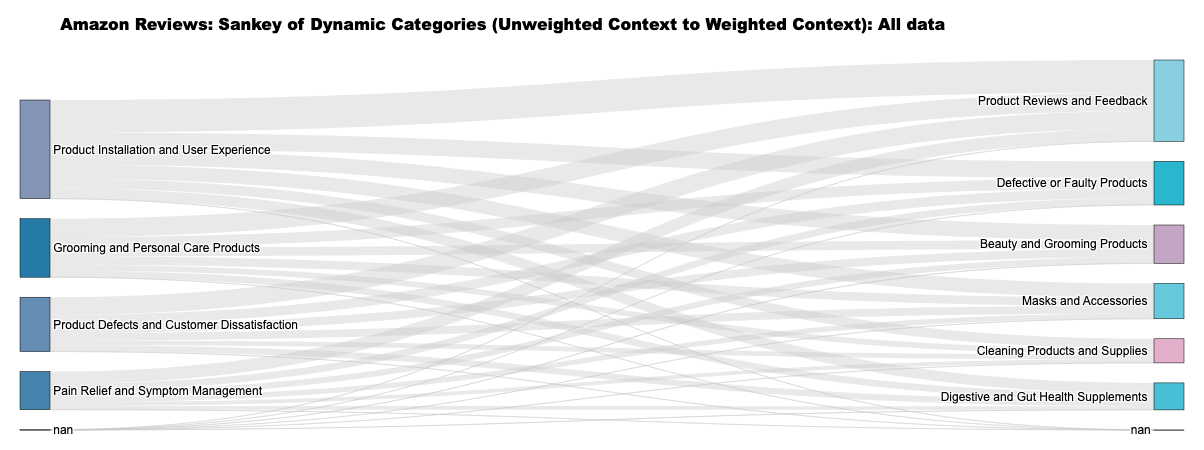}
        
    \end{subfigure}

    \caption{Detailed Sankey diagrams showing cluster transitions for Amazon Product Reviews.}
\end{figure}

\begin{table}[ht!]
    \centering
    \caption{Distribution of Review Categories across Experimental Scenarios (Amazon Product Reviews)}
    \label{tab:amazon_distribution_full}
    \normalsize 
    \renewcommand{\arraystretch}{1.3} 
    
    \begin{tabularx}{\linewidth}{@{} l >{\raggedright\arraybackslash}X r r r @{}}
        \toprule
        \textbf{Scenario} & \textbf{Category-Cluster Titles} & \textbf{All Data} & \textbf{\makecell[r]{W/o\\Irrelevant}} & \textbf{\makecell[r]{W/o Irrelevant\\\& Outliers}} \\
        \midrule
        
        \multirow{7}{*}{\makecell[l]{No context \\ (Baseline)}} 
        & Assistive Devices and Aids & 18,105 & 17,712 & 13,821 \\
        & Grooming and Personal Care Products & 36,451 & 35,309 & 27,846 \\
        & Cleaning and Maintenance Products & 11,626 & 11,241 & 9,121 \\
        & Pain and Symptom Relief & 15,788 & 15,110 & 10,226 \\
        & Personal Grooming and Hygiene & 33,066 & 32,526 & 26,922 \\
        & Product Installation and User Experience & 61,491 & 58,699 & 46,492 \\
        & Product Quality and Performance Issues & 39,244 & 38,105 & 30,391 \\
        \midrule \addlinespace[2pt]
        
        \multirow{5}{*}{\makecell[l]{Unweighted \\ Context (SSAS)}}
        & Beauty and Grooming Products & 23,927 & 23,492 & 19,245 \\
        & Pain Relief and Symptom Management & 23,764 & 23,022 & 16,749 \\
        & Product Defects and Customer Dissat. & 33,853 & 32,670 & 24,893 \\
        & Positive Experiences and Reactions & 19,570 & 18,143 & 14,116 \\
        & Product Functionality and Performance & 18,060 & 16,763 & 11,342 \\
        \midrule \addlinespace[2pt]
        
        \multirow{5}{*}{\makecell[l]{Weighted \\ Context (wSSAS)}}
        & Cleaning Products and Supplies & 15,035 & 14,508 & 11,668 \\
        & Defective or Faulty Products & 27,177 & 26,134 & 20,044 \\
        & Digestive and Gut Health Supplements & 16,820 & 16,278 & 11846 \\
        & Masks and Accessories & 22,032 & 21,565 & 16,417 \\
        & Product Reviews and Feedback & 50,694 & 47,801 & 36,837 \\
        \addlinespace[4pt] 
        \bottomrule
    \end{tabularx}
\end{table}

\clearpage
Goodreads Book Reviews (All Data)

\begin{figure}[ht!]
    \centering
    
    \begin{subfigure}{\textwidth}
    \caption{No Context (Baseline) vs Weighted Context Summary (wSSAS)}
        \centering
        \includegraphics[trim={0 0 0 35pt}, clip, width=0.9\linewidth]{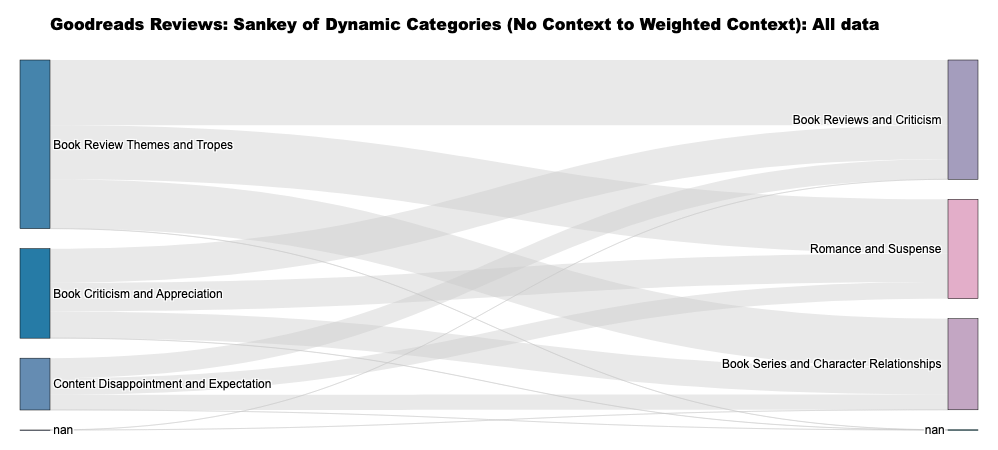}
        
    \end{subfigure}
    
    \vspace{10pt}

    \begin{subfigure}{\textwidth}
    \caption{No Context (Baseline) vs Unweighted Context Summary (SSAS)}
        \centering
        \includegraphics[trim={0 0 0 35pt}, clip, width=0.9\linewidth]{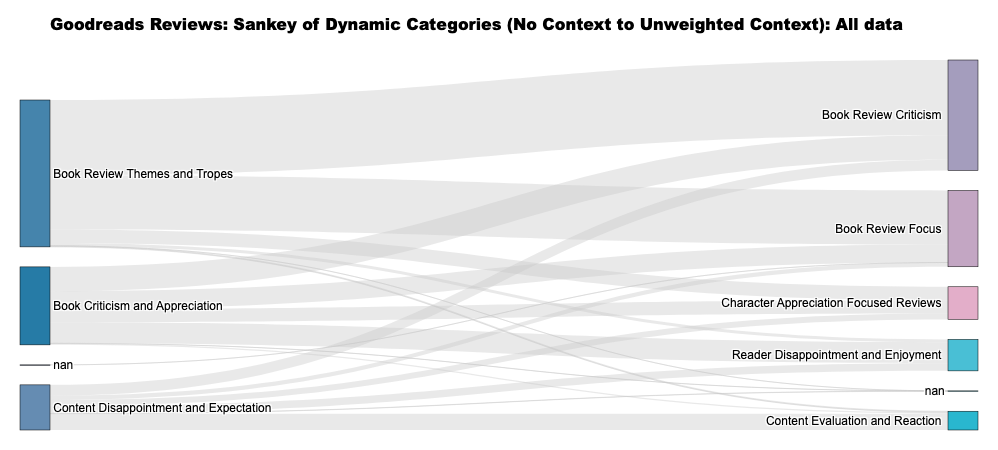}
        
    \end{subfigure}

    \vspace{10pt}

    \begin{subfigure}{\textwidth}
    \caption{Unweighted Context Summary (SSAS) vs Weighted Context Summary (wSSAS)}
        \centering
        \includegraphics[trim={0 0 0 35pt}, clip, width=0.9\linewidth]{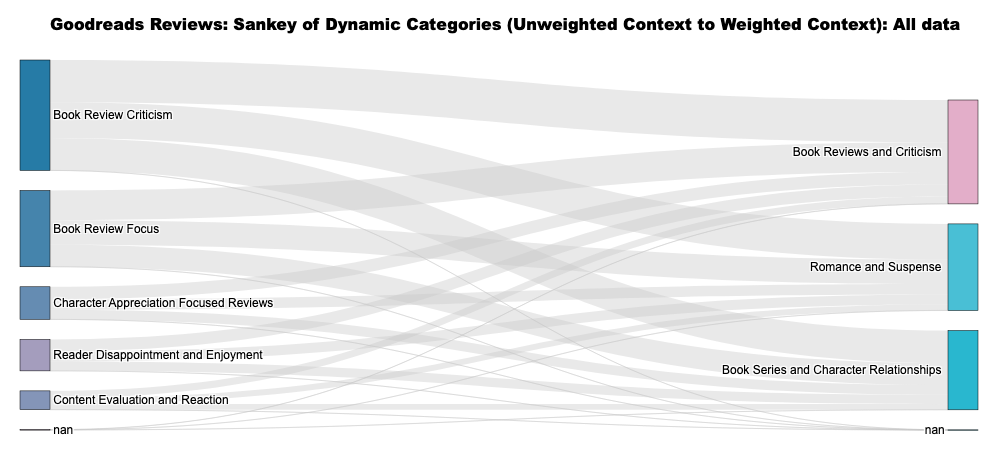}
        
    \end{subfigure}

    \caption{Detailed Sankey diagrams showing cluster transitions for Goodreads Book Reviews.}
\end{figure}

\begin{table}[ht!]
    \centering
    \caption{Distribution of Review Categories across Experimental Scenarios (Goodreads Book Reviews)}
    \label{tab:goodreads_distribution_full}
    \normalsize 
    \renewcommand{\arraystretch}{1.3} 
    
    \begin{tabularx}{\linewidth}{@{} l >{\raggedright\arraybackslash}X r r r @{}}
        \toprule
        \textbf{Scenario} & \textbf{Category-Cluster Titles} & \textbf{All Data} & \textbf{\makecell[r]{W/o\\Irrelevant}} & \textbf{\makecell[r]{W/o Irrelevant\\\& Outliers}} \\
        \midrule 
        
        \multirow{3}{*}{\makecell[l]{No context \\ (Baseline)}} 
        & Book Criticism and Appreciation & 45,460 & 40,413 & 31,147 \\
        & Book Review Themes and Tropes & 85,634 & 82,708 & 73,520 \\
        & Content Disappointment and Expectation & 26,311 & 20,277 & 12,464 \\

        \midrule \addlinespace[2pt]
        
        \multirow{5}{*}{\makecell[l]{Unweighted \\ Context (SSAS)}}
        & Book Review Criticism & 64,388 & 60,586 & 52,847 \\
        & Book Review Focus & 44,495 & 43,416 & 36,769 \\
        & Character Appreciation Focused Reviews & 19,023 & 16,659 & 13,108 \\
        & Content Evaluation and Reaction & 10,946 & 7,889 & 4,000 \\
        & Reader Disappointment and Enjoyment & 18,347 & 14,761 & 10,350 \\

        \midrule \addlinespace[2pt]
        
        \multirow{3}{*}{\makecell[l]{Weighted \\ Context (wSSAS)}}
        & Book Reviews and Criticism & 60,574 & 55,734 & 46,506 \\
        & Book Series and Character Relationships & 46,246 & 38,511 & 27,622 \\
        & Romance and Suspense & 50,483 & 49,057 & 42,912 \\

        \addlinespace[4pt] 
        \bottomrule
    \end{tabularx}
\end{table}

\clearpage
\section{Technical Stack and Implementation Environment}
\label{app:tech_stack}

This section details the software, models, and mathematical frameworks utilized to implement and validate the wSSAS methodology.

\subsection{Large Language Models (LLMs)}
\begin{itemize}
    \item \textbf{Primary Inference Engine:} \texttt{Gemini 2.0 Flash Lite} was utilized for hierarchical text summarization and categorization across Themes, Stories, and Clusters.
    \item \textbf{LLM-as-a-Judge:} The same model facilitated the reference-free evaluation framework, performing Question-Answer Generation (QAG) and qualitative G-Eval scoring.
\end{itemize}

\subsection{Embedding Models and Vectorization}
\begin{itemize}
    \item \textbf{High-Dimensional Vectorization:} \texttt{text-embedding-005} was employed to generate high-fidelity vector representations of the raw text, leveraging its advanced semantic grasp for processing heterogeneous datasets.
    \item \textbf{Semantic Similarity Engine:} The \texttt{sentence-transformers/all-MiniLM-L6-v2} model was utilized within the QAG framework to calculate cosine similarity between true and extracted responses.
\end{itemize}

\subsection{Internal Validation Metrics}
Cluster integrity was mathematically confirmed using three primary metrics to ensure the cohesion of the generated themes:

\begin{enumerate}
    \item \textbf{Silhouette Score:} Measures internal cohesion ($a$) versus cluster separation ($b$). 
    \item \textbf{Davies-Bouldin Index:} Measures the average similarity between clusters. Lower scores indicate better separation between thematic groups.
    \item \textbf{Calinski-Harabasz (CH) Index:} Evaluates the ratio of between-cluster dispersion to within-cluster dispersion (the Variance Ratio Criterion).
\end{enumerate}

\end{appendices}

\end{document}